\documentclass[runningheads]{llncs}

\usepackage{eccv}

\usepackage{eccvabbrv}

\usepackage{graphicx}
\usepackage{booktabs}

\usepackage[accsupp]{axessibility}  

\usepackage[pagebackref,breaklinks,colorlinks,citecolor=eccvblue]{hyperref}

\usepackage{orcidlink}

\usepackage{multirow,booktabs,pifont}

\DeclareMathOperator*{\argmax}{arg\,max}

\begin{document}

\title{Discovering Novel Actions from Open World Egocentric Videos with Object-Grounded Visual Commonsense Reasoning} 

\titlerunning{Abbreviated paper title}

\author{Sanjoy Kundu\inst{1} \and
Shubham Trehan\inst{1} \and
Sathyanarayanan N. Aakur\inst{1}}

\authorrunning{S. Kundu et al.}

\institute{CSSE Department, Auburn University,\\
Auburn, AL, USA 36849\\
\email{\{szk0266,szt0113,san0028\}@auburn.edu}}

\maketitle

\begin{abstract}
Learning to infer labels in an open world, i.e., in an environment where the target ``labels'' are unknown, is an important characteristic for achieving autonomy. Foundation models, pre-trained on enormous amounts of data, have shown remarkable generalization skills through prompting, particularly in zero-shot inference. 
However, their performance is restricted to the correctness of the target label's search space, i.e., candidate labels provided in the prompt. 
This target search space can be unknown or exceptionally large in an open world, severely restricting their performance. To tackle this challenging problem, we propose a two-step, neuro-symbolic framework called ALGO - \underline{A}ction \underline{L}earning with \underline{G}rounded \underline{O}bject recognition that uses symbolic knowledge stored in large-scale knowledge bases to infer activities in egocentric videos with limited supervision. First, we propose a neuro-symbolic prompting approach that uses \textit{object-centric} vision-language models as a noisy oracle to ground objects in the video through evidence-based reasoning. Second, driven by prior commonsense knowledge, we discover plausible activities through an energy-based symbolic pattern theory framework and learn to ground knowledge-based action (verb) concepts in the video. Extensive experiments on four publicly available datasets (EPIC-Kitchens, GTEA Gaze, GTEA Gaze Plus, and Charades-Ego) demonstrate its performance on open-world activity inference. We also show that ALGO can be extended to zero-shot inference and demonstrate its competitive performance on the Charades-Ego dataset. 
\end{abstract}

\section{Introduction}
\label{sec:intro}
Humans display a remarkable ability to recognize unseen concepts (actions, objects, etc.) by associating known concepts gained through prior experience and reasoning over their attributes. Key to this ability is the notion of ``grounded'' reasoning, where abstract concepts can be mapped to the perceived sensory signals to provide evidence to confirm or reject hypotheses. In this work, we aim to create a computational framework that tackles open-world egocentric activity understanding. We define an activity as a complex structure whose semantics are expressed by a combination of actions (verbs) and objects (nouns). To recognize an activity, one must be cognizant of the object label, action label, and the possibility of any combination since not all actions are plausible for an object.
Supervised learning approaches~\cite{wang2013action,sigurdsson2018actor,ma2016going,dosovitskiyimage} have been the dominant approach to activity understanding but are trained in a ``closed'' world, where there is an implicit assumption about the target labels. The videos during inference will always belong to the label space seen during training.
Zero-shot learning approaches~\cite{zhang2017first,lin2022egocentric,zhao2023lavila,ashutosh2023hiervl} relax this assumption by considering disjoint ``seen'' and ``unseen'' label spaces where all labels are not necessarily represented in the training data. This setup is a \textit{known} world, where the target labels are pre-defined and aware during training.
In this work, we define an \textit{open} world to be one where the target labels are unknown during both training and inference. The goal is to recognize elementary concepts and infer the activity.

\begin{figure*}[t]
    \centering
    \includegraphics[width=0.9\textwidth]{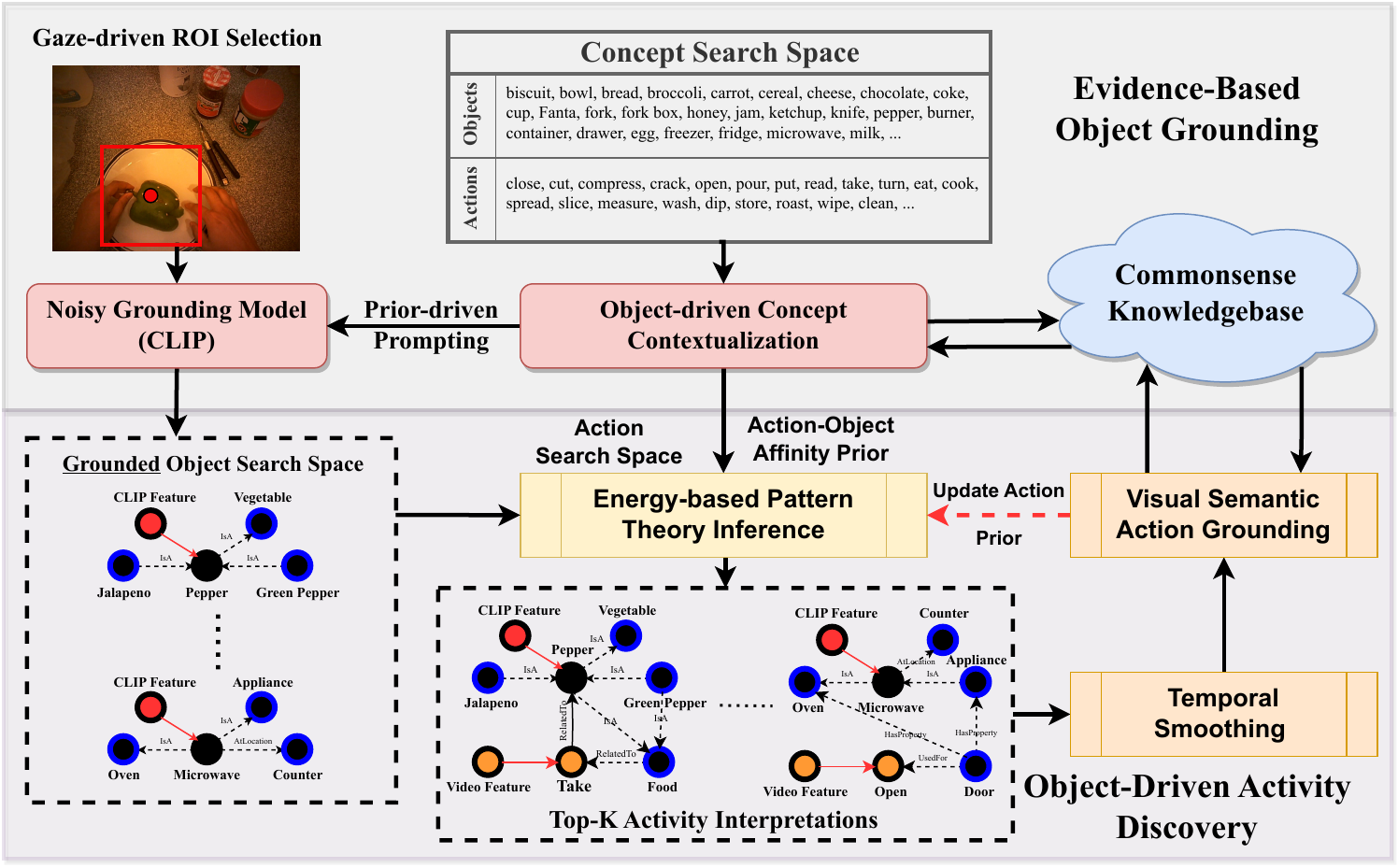}
    \caption{\textbf{Overall architecture} of the proposed approach (ALGO) is illustrated here. Using a two-step process, we first \textit{ground} the objects within a gaze-driven ROI using CLIP~\cite{radford2021learning} as a noisy oracle before reasoning over the plausible activities performed in the video. The inferred activity and action (verb) are grounded in prior knowledge and visual features to refine the activity interpretations. }
    \label{fig:arch}
\end{figure*}

\textit{Foundation models}~\cite{bommasani2021opportunities}, pre-trained on large amounts of data, have shown tremendous performance on different problems such as question answering~\cite{devlin2018bert}, zero-shot object recognition~\cite{radford2021learning}, and action recognition~\cite{lin2022egocentric}. 
Self-supervised pre-training~\cite{zhao2023lavila,dosovitskiyimage} has helped improve their generalization. 
However, their ability to perform open-world inference is constrained by two factors. First, the search space (i.e., target label candidates) must be well-defined since their output is constrained to what is presented to them (or ``prompted''), which requires prior knowledge about the environment. Second, their performance is dependent on the span of their pre-training data. Models trained on third-person views may not generalize to egocentric videos due to the limited capability to \textit{ground} semantics in visual data and \textit{reason} over object affordances.
Learning these associations during pre-training is challenging since it requires data from every possible combination of concepts. 
%
We propose to tackle this problem using a neuro-symbolic framework that leverages advances in multimodal foundation models to ground concepts from symbolic knowledge bases, such as ConceptNet~\cite{speer2017conceptnet}, in visual data. The overall approach is shown in Figure~\ref{fig:arch}. Using the energy-based pattern theory formalism~\cite{aakur2022knowledge,aakur2019generating,grenander1996elements} to represent symbolic knowledge, we ground objects (nouns) using CLIP~\cite{radford2021learning} as a noisy oracle. Driven by prior knowledge, novel activities (verb+noun) are inferred, and the associated action (verb) is grounded in the video to learn visual-semantic associations for novel, unseen actions.

The \textbf{contributions} of this work are three-fold: (i) We present a neuro-symbolic framework to leverage compositional properties of objects to prompt CLIP for evidence-based grounding. (ii) We propose object-driven activity discovery as a mechanism to reason over prior knowledge and provide action-object affinities to constrain the search space. (iii) We demonstrate that the inferred activities can be used to ground unseen actions (verbs) from symbolic knowledge in egocentric videos, which can generalize to unseen and unknown action spaces.

\section{Related Works}
\textbf{Egocentric video analysis} has been extensively explored in computer vision literature, having applications in virtual reality~\cite{han2020megatrack} and human-machine interaction. Various tasks have been proposed, such as question-answering~\cite{fan2019egovqa}, summarization~\cite{lu2013story}, gaze prediction~\cite{li2013learning,fathi2012learning,aakur2021unsupervised}, and action recognition~\cite{li2019deep}, among others. Success has been driven by the development of large-scale datasets such as Ego-4D~\cite{grauman2022ego4d}, Charade-Ego~\cite{sigurdsson2018actor}, GTEA Gaze~\cite{fathi2012learning}, GTEA Gaze Plus~\cite{li2013learning}, and EPIC-Kitchens~\cite{damen2020epic}. 
In the context of egocentric activity recognition, which is the focus of this work, supervised learning has been the predominant approach. Researchers have explored various techniques, such as modeling spatial-temporal dynamics~\cite{Sudhakaran_2019_CVPR}, using appearance and motion cues for recognition~\cite{ma2016going}, hand-object interaction~\cite{Zhou_2016_CVPR,Wang_2021_ICCV}, and time series modeling of motion information~\cite{Ryoo_2015_CVPR}, to name a few. 
Some studies have addressed the data-intensive nature by exploring zero-shot learning~\cite{zhang2017first,sigurdsson2018actor}.
KGL~\cite{aakur2022knowledge} is one of the first works to address the problem of \textbf{open-world understanding}. They represent knowledge elements derived from ConceptNet~\cite{speer2017conceptnet}, using pattern theory~\cite{aakur2019generating,desouza2016pattern,grenander1996elements}. However, their method relies on an object detector to ground objects in a source domain before mapping concepts to the target space using ConceptNet-based semantic correspondences. This approach has limitations: (i) false alarms may occur when the initial object detector fails to detect the object of interest, leading to the use of the \textit{closest} object to the gaze, and (ii) reliance on ConceptNet for correspondences from the source domain to the target domain, resulting in objects being disregarded if corresponding probabilities are zero. Other efforts in \textbf{open-world learning} have primarily focused on \textit{object-centric} tasks, such as open-world object detection~\cite{Gu2021OpenvocabularyOD, du2022learning, dong2022open}, which do not address the combinatorial problems inherent in open-world \textit{activity} recognition.

\textbf{Vision-language modeling} has gained significant attention in the community, driven by the success of transformer models~\cite{vaswani2017attention} in natural language processing, such as BERT~\cite{devlin2018bert}, RoBERTa~\cite{liu2019roberta}, OpenAI's GPT series~\cite{radford2018improving,radford2019language,brown2020language,bubeck2023sparks}, and ELECTRA~\cite{clark2020electra}. 
The development of object-centric foundation models has enabled impressive capabilities in zero-shot object recognition in images, as demonstrated by CLIP~\cite{radford2021learning}, DeCLIP~\cite{li2022supervision}, and ALIGN~\cite{jia2021scaling}. 
These approaches rely on large amounts of image-text pairs, often in the order of \textit{billions}, to learn visual-semantic representations using various forms of contrastive learning~\cite{khosla2020supervised,chen2020simple}. 
Recent works, such as EGO-VLP~\cite{lin2022egocentric}, Hier-VL~\cite{ashutosh2023hiervl}, LAVILLA~\cite{zhao2023lavila}, and CoCa~\cite{yu2022coca} have expanded the scope of multimodal foundation models to include egocentric videos and have achieved impressive performance in zero-shot generalization. However, these approaches require substantial amounts of curated pre-training data to learn semantic associations among concepts for egocentric activity recognition. \textbf{Neuro-symbolic models}~\cite{nye2021improving,NEURIPS2020_94c28dcf,NEURIPS2022_3ff48dde,aakur2022knowledge} show promise in reducing the increasing dependency on data. Our approach extends the idea of neuro-symbolic reasoning to address egocentric, open-world activity recognition.

\section{Proposed Framework: ALGO}
\textbf{Problem Formulation.} We address the task of recognizing unknown activities in egocentric videos within an open-world setting. Our objective is to develop a framework that can learn to identify elementary concepts, establish semantic associations, and systematically explore, evaluate, and reject combinations to arrive at an interpretation that best describes the observed activity class. In this context, we define the target classes as activities, which are composed of elementary concepts such as actions (verbs) and objects (nouns). These activities are formed by combining concepts from two distinct sets: an object search space ($G_{obj}$) and an action search space ($G_{act}$). These sets define the pool of available elementary concepts (objects and actions) that can be used to form an activity (referred to as the ``target label''). 

\textbf{Overview.} We propose ALGO (Action Learning with Grounded Object recognition), illustrated in Figure~\ref{fig:arch}, to tackle the problem of discovering novel actions in an open world. 
Given a search space of elementary concepts, we first hypothesize the presence of plausible objects through evidence-based object grounding (Section~\ref{sec:NSG}) by exploring prior knowledge from a symbolic knowledge base. A noisy grounding model provides visual grounding to generate a grounded object search space. We then use an energy-based inference mechanism (Section~\ref{sec:inference}) to discover the plausible actions that can be performed on the ground object space, driven by prior knowledge from symbolic knowledge bases, to recognize unseen and unknown activities (action-object combinations) without supervision. A visual-semantic action grounding mechanism (Section~\ref{sec:temporalGround}) then provides feedback to ground semantic concepts with video-based evidence for discovering composite activities without explicit supervision. Although our framework is flexible to work with any noisy grounding model and knowledge base, we use CLIP~\cite{radford2021learning} and ConceptNet~\cite{speer2017conceptnet}, respectively. 

\textbf{Knowledge Representation.} We use Grenander's pattern theory formalism~\cite{grenander1996elements} to represent the knowledge elements and build a contextualized activity interpretation that integrates neural and symbolic elements in a unified, energy-based representation. 
Pattern theory provides a flexible framework to help reason over variables with varying underlying dependency structures by representing them as compositions of simpler patterns. 
These structures, called configurations, are composed of atomic elements called \textit{generators} ($\{g_1, g_2, \ldots g_i\} \in G_s$), which connect through local connections called \textit{bonds} ($\{\beta_1, \beta_2, \ldots \beta_i\} \in g_i$). The collection of all generators is called the \textit{generator space} ($G_s$), with each generator possessing an arbitrary set of bonds, defined by its \textit{arity}. Bonds between generators are constrained through local and global \textit{regularities}, as defined by an overarching graph structure. A probability structure over the representations captures the diversity of patterns. 
We refer the reader to Aakur \textit{et al.}~\cite{aakur2019generating} and de Souza \textit{et al.} ~\cite{desouza2016pattern} for a deeper exploration of pattern theory.

\subsection{Evidence-based Object Grounding with Prior-driven Prompting}\label{sec:NSG}
The first step in our framework is to assess the plausibility of each object concept (represented as generators $\{g^o_1, g^o_2, \ldots g^o_i\} \in G_{obj}$) by \textit{grounding} them in the input video \textit{$V_i$}. 
We define \textit{grounding} as gathering evidence from the input data to support a concept's presence (or absence) in the final interpretation. While object-centric vision-language foundation models such as CLIP~\cite{radford2021learning} have shown impressive abilities in zero-shot object recognition in images, egocentric videos provide additional challenges such as camera motion, lens distortion, and out-of-distribution object labels. Follow-up work~\cite{menon2022visual} has focused on addressing them to a certain extent by probing CLIP for explainable object classification. However, they do not consider \textit{compositional} properties of objects and alternative labels for verifying their presence in the video. To address this issue, we propose a neuro-symbolic \textit{evidence-based} object grounding mechanism to compute the likelihood of an object in a given frame. For each object generator ($g^o_i$) in the search space ($G_{obj}$), we first compute a set of compositional \textit{ungrounded} generators by constructing an ego-graph of each object label ($E_{{g}^o_i}$) from ConceptNet~\cite{speer2017conceptnet} and limiting edges to those that express \textit{compositional} properties such as \texttt{IsA, UsedFor, HasProperty} and \texttt{SynonymOf}. 
Given this set of \textit{ungrounded} generators ($\{\bar{g}^o_i\} \forall g^o_i \in G_{obj}$), we then prompt CLIP to provide likelihoods for each ungrounded generator $p(\bar{g}^o_i \vert I_t)$ to compute the \textit{evidence-based} likelihood for each \textit{grounded} object generator $\underline{g}^o_i$ as defined by the probability 
\begin{align}
    p(\underline{g}^o_i \vert \bar{g}^o_i, I_t, K_{CS}) = p(\underline{g}^o_i | I_t)*\left\lVert \sum_{\forall \bar{g}^o_i} 
    p({g}^o_i, \bar{g}^o_i \vert E_{{g}^o_i}) * p(\bar{g}^o_i) | I_t)    \right\rVert^2
    \label{eqn:groundingProb}
\end{align}
where $p({g}^o_i, \bar{g}^o_i \vert E_{{g}^o_i})$ is the edge weight from the edge graph $E_{{g}^o_i}$ (sampled from a knowledge graph $K_{CS}$) that acts as a prior for each ungrounded evidence generator $\bar{g}^o_i$ and $p(\bar{g}^o_i | I_t)$ is the likelihood from CLIP for its presence in each frame $I_t$. Hence the probability of the presence of a \textit{grounded} object generator is determined by (i) its image-based likelihood, (ii) the image-based likelihood of its evidence generators, and (iii) support from prior knowledge for the presence of each evidence generator. Hence, we ground the object generators in each video frame by constructing and evaluating the evidence to support each grounding assertion and provide an interpretable interface to video object grounding. 
Empirically, in Section~\ref{sec:ablation}, we see that this evidence-based grounding outperforms n\"aive CLIP-based grounding. 
To navigate clutter and focus only on the object involved in the activity (i.e., the packaging problem~\cite{maguire2008speaking}), we use human gaze to select a $200\times200$ region centered around the gaze position (from the human user if available, else we approximate it with center bias~\cite{li2013learning}).

\subsection{Object-driven Activity Discovery}\label{sec:inference}
The second step in our approach is to discover plausible activities performed in the given video. 
We take an object affordance-based approach to activity inference, constraining the activity label (verb+noun) to those that conform to affordances defined in prior knowledge. 
We first construct an ``\textit{action-object affinity}'' function that provides a \textit{prior} probability for the validity of an activity. 
The probability of each action-object combination is computed by taking a weighted sum of the edge weights (direct and indirect) along each path that connects them in ConceptNet.
An exponential decay function is applied to each term to avoid generating excessively long paths that can introduce noise into the reasoning process. 
Finally, we filter out paths that do \textit{not} contain compositional assertions (\texttt{UsedFor, HasProperty, IsA}) since generic assertions (such as \texttt{RelatedTo}) do not explicitly capture object affordances. 
The probability of an activity (defined by an action generator $g^a_i$ and a grounded object generator $\underline{g}^o_j$) is given by
\begin{equation}
    p(g^a_i, \underline{g}^o_j \vert K_{CS}) = \argmax_{\forall E \in K_{CS}} \sum_{(\bar{g}_{m}, \bar{g}_{n}) \in E}{w_k * K_{CS}(\bar{g}_{m}, \bar{g}_{n})}
    \label{eqn:affinity}
\end{equation}
where $E$ is the collection of all paths between $g^a_i$ and $\underline{g}^o_j$ in a commonsense knowledge graph $K_{CS}$, $w_k$ is a weight drawn from an exponential decay function based on the distance of the node $\underline{g}^o_j$ from $g^a_i$. After filtering for compositional properties, the path with the maximum weight is chosen with the optimal action-object affinity. The process is repeated for all activities in the search space. 

\textbf{Energy-based Activity Inference.} To reason over the different activity combinations, we assign an energy term to each activity label, represented as a \textit{configuration}.
These are complex structures composed of individual generators that combine through bonds dictated by their affinity functions. In our case, each activity interpretation is a configuration composed of a grounded object generator ($\underline{g}^o_i$), its associated ungrounded evidence generators ($\bar{g}^o_j$), an action generator ($g^a_k$) and ungrounded generators from their affinity function, connected via an underlying graph structure. This graph structure will vary for each configuration depending on the presence of affinity-based bonds derived from ConceptNet. Hence, the \textit{energy} of a configuration $c_i$ is given by 
\begin{equation}
\begin{split}
    E(c) &= \phi(p(\underline{g}^o_i \vert \bar{g}^o_j, I_t, K_{CS})) + \phi(p(g^a_k, \underline{g}^o_i \vert K_{CS})) + \phi(p(g^a_k | I_t))
\end{split}
    \label{eqn:configEnergy}
\end{equation}
where the first term provides the energy of grounded object generators (from Equation~\ref{eqn:groundingProb}), the second term provides the energy from the affordance-based affinity between the action and object generators (from Equation~\ref{eqn:affinity}), and the third term is the likelihood of an action generator. The probability of a configuration $c$ is given by $p(c) \propto exp(-E(c))$. Hence, the lower the energy, the higher its probability. We initially set $\phi(p(g^a_k | I_t)) = 1$ to reason over all possible actions for each object and later update this using a posterior refinement process (Section~\ref{sec:posterior}). Hence, activity inference becomes an optimization over Equation~\ref{eqn:configEnergy} to find the configuration (or activity interpretation) with the least energy. For tractable computation, we use the MCMC-based simulated annealing mechanism proposed in KGL~\cite{aakur2022knowledge} to avoid an expensive brute-force search over all verb-noun combinations. 
If action priors are available from video-centric foundation models~\cite{lin2022egocentric,zhao2023lavila}, $\phi(p(g^a_k | I_t))$ can be initialized by prompting it with plausible action labels. Empirically, in Section~\ref{sec:results}, we can show that leverage vision-language foundation models, if available, to significantly improve the performance. 

\subsection{Visual-Semantic Action Grounding}\label{sec:temporalGround}
The third step in our framework is the idea of visual-semantic action grounding, where we aim to learn to ground the inferred actions (verbs) from the overall activity interpretation. While CLIP provides a general purpose, if noisy, object grounding method, a comparable approach for actions does not exist. Hence, we learn an action grounding model by bootstrapping a simple function ($\psi(g^a_i, f_V)$) to map clip-level visual features to the semantic embedding space associated with ConceptNet, called ConceptNet Numberbatch~\cite{speer2017conceptnet}. The mapping function is a simple linear projection to go from the symbolic generator space ($g^a_i \in G_{act}$) to the semantic space ($f^a_i$), which is a 300-dimension ($\mathbb{R}^{1\times300}$) vector representation explicitly trained to capture concept-level attributes captured in ConceptNet. While there can be many sophisticated mechanisms~\cite{lin2022egocentric,ashutosh2023hiervl}, including contrastive loss-based training, we use the mean squared error (MSE) loss as the objective function to train the mapping function since our goal is to provide a mechanism to ground abstract concepts from the knowledge-base in the video data. We leave the exploration of more sophisticated grounding mechanisms to future work.

\textbf{Temporal Smoothing} Since we predict frame-level activity interpretations to account for gaze transitions, we first perform temporal smoothing to label the entire video clip before training the mapping function $\psi(g^a_i, f_V)$ to reduce noise in the learning process. For each frame in the video clip, we take the five most common actions predicted at the \textit{activity} level (considering the top-10 predictions) and sum their energies to consolidate activity predictions and account for erroneous predictions. We then repeat the process for the entire clip, i.e., get the top-5 actions based on their frequency of occurrence at the frame level and consolidated energies across frames. These five actions provide targets for the mapping function $\psi(g^a_i, f_V)$, which is then trained with the MSE function. We use the top-5 action labels as targets to limit the effect of frequency bias. 

\textbf{\textbf{Posterior-based Activity Refinement.}}\label{sec:posterior}
The final step in our framework is an iterative refinement process that updates the action concept priors (the third term in Equation~\ref{eqn:configEnergy}) based on the predictions of the visual-semantic grounding mechanism described in Section~\ref{sec:temporalGround}. Since our predictions are made on a per-frame basis, it does not consider the overall temporal coherence and visual dynamics of the clip. Hence, there can be contradicting predictions for the actions done over time. 
Hence, we iteratively update the action priors for the energy computation to re-rank the interpretations based on the clip-level visual dynamics. 
We iteratively refine the activity labels and update the visual-semantic action grounding modules simultaneously by alternating between posterior update and action grounding until the generalization error (i.e., the performance on unseen actions) saturates, which indicates overfitting. 

\textbf{Implementation Details.} 
We use an S3D-G network pre-trained by Miech \textit{et al.}~\cite{miech2019howto100m,miech2020end} on Howto100M~\cite{miech2019howto100m} as our visual feature extraction for visual-semantic action grounding. We use a CLIP model with the ViT-B/32~\cite{dosovitskiyimage} as its backbone network. ConceptNet was used as our source of commonsense  knowledge for neuro-symbolic reasoning, and ConceptNet Numberbatch~\cite{speer2017conceptnet} was used as the semantic representation for action grounding. 
The mapping function, defined in Section~\ref{sec:temporalGround}, was a 1-layer feedforward network trained with the MSE loss for 100 epochs with a batch size of 256 and learning rate of $10^{-3}$. 
Generalization errors were used to pick the best model. 
Experiments were conducted on a desktop with a 32-core AMD ThreadRipper and an NVIDIA TitanRTX.

\section{Experimental Evaluation}\label{sec:results}

\textbf{Data.} To evaluate the open-world inference capabilities, we evaluate the approach on GTEA Gaze~\cite{fathi2012learning}, GTEA GazePlus~\cite{li2013learning}, and EPIC-Kitchens-100~\cite{Damen2021PAMI,Damen2022RESCALING} datasets, which contain egocentric, multi-subject videos of meal preparation activities. 
GTEA Gaze and GazePlus have frame-level gaze information and activity labels, providing an ideal test bed for our setup. 
EPIC-Kitchens-100 is a much larger dataset and does not have gaze information, offering a much more challenging evaluation of the approach. 
We also evaluate on Charades-Ego~\cite{sigurdsson2018actor}, a larger egocentric video dataset focused on activities of daily living, to evaluate on the zero-shot setting. 
The evaluation datasets, under an open-world setting, offer a significant challenge with an increasingly larger search space. The GTEA Gaze dataset has 10 verbs and 38 nouns (search space of 380 activities), while GTEA GazePlus has 15 verbs and 27 nouns (search space of 405), Charades-Ego has 33 verbs and 38 nouns (search space of 1254), and Epic-Kitchens has 97 verbs and 300 nouns (search space of 29100).

\begin{table*}[t]
    \centering
    \resizebox{0.99\textwidth}{!}{
    \begin{tabular}{|c|c|c|c|c|c|c|c|c|}
    \toprule
    \multirow{2}{*}{\textbf{Approach}}& \multirow{2}{*}{\textbf{Search }} & \multirow{2}{*}{\textbf{VLM?}} & \multicolumn{3}{|c|}{\textbf{GTEA Gaze}} & \multicolumn{3}{|c|}{\textbf{GTEA GazePlus}}\\
    \cmidrule{4-9}
    & \textbf{Space} &  & \textbf{Object} & \textbf{Action} & \textbf{Activity} & \textbf{Object} & \textbf{Action} & \textbf{Activity}\\
    \toprule
    Two-Stream CNN~\cite{simonyan2014two} & Closed & \ding{55} & 38.05 & 59.54 & \underline{53.08} & \underline{61.87} & 58.65 & 44.89\\
    IDT~\cite{wang2013action} & Closed & \ding{55} & \underline{45.07} & \underline{75.55} & 40.41 & 53.45 & \underline{66.74} & \underline{51.26}\\
    Action Decomposition~\cite{zhang2017first} & Closed & \ding{55} & \textbf{60.01} & \textbf{79.39} & \textbf{55.67} & \textbf{65.62} & \textbf{75.07} & \textbf{57.79}\\
    \midrule
    Random & Known & \ding{55} & 3.22 & 7.69 & 2.50 & 3.70 & 4.55 & 2.28\\
    Action Decomposition ZSL*~\cite{zhang2017first} & Known & \ding{55} & \textbf{65.81} & \textbf{89.17} & \textbf{68.70} & \textbf{53.40} & \textbf{32.48} & \textbf{29.19}\\
    ALGO ZSL (Ours) & Known & \ding{55} & \underline{49.47} & \underline{74.74} & \underline{27.34} & \underline{47.67} & \underline{29.31} & \underline{16.68}\\
    \midrule
    KGL~\cite{aakur2022knowledge} & Open & \ding{55} &  5.12 & 8.04 & 4.91 & 14.78 & 6.73 & 10.87\\
    KGL+CLIP~\cite{aakur2022knowledge} & Open & \ding{55} &  {10.36} & {8.15} & {9.21} & {20.49} & {9.23} & {14.86}\\
    ALGO (Ours) & Open & \ding{55} &  \textbf{13.07} & \textbf{17.05} & \textbf{15.05} & \textbf{26.23} & \textbf{11.44} & \textbf{18.84}\\
    \midrule
    EgoVLP~\cite{lin2022egocentric} & Open & \ding{51} &  10.17 & 8.45 & 9.31 & 29.43 & 17.17 & 23.30 \\
    LaViLa~\cite{zhao2023lavila} & Open & \ding{51} &  6.07 & 23.07 & 14.57 & 28.27 & 25.47 & 26.87 \\
    ALGO+LaViLa & Open & \ding{51} &  \textbf{17.50} & \textbf{26.60} & \textbf{22.05} & \textbf{30.74} & \textbf{27.00} & \textbf{28.87}\\
    \bottomrule
    \end{tabular}
    }
    \caption{\textbf{Open-world activity recognition} performance on the GTEA Gaze and GTEA Gaze Plus datasets. We compare approaches with a closed search space, those with a known search space, and those with a partially open one. Accuracy is reported for predicted objects, actions, and activities. VLM: Vision-Language Model pre-trained on egocentric video data. * indicates training on ``seen'' classes from the same dataset(s) and leave-one-action-out evaluation.}
    \label{tab:gaze_results}
    \vspace{-5mm}
\end{table*}

\textbf{Evaluation Metrics.} Following prior work in open-world activity recognition~\cite{aakur2022knowledge,aakur2019generating}, we use accuracy to evaluate action and object recognition and use \textit{word-level} accuracy for evaluating the activity (verb+noun) recognition performance. It provides a less-constrained measurement to measure the quality of predictions beyond accuracy by considering all units without distinguishing between insertions, deletions, or misclassifications. 
This helps quantify the performance without penalizing semantically similar interpretations. We use class-wise mAP to evaluate ALGO in the zero-shot learning setup on Charades-Ego.

\textbf{Baselines.} We compare against different egocentric action recognition approaches, including those with a closed-world learning setup. For open-world inference, we compare it against Knowledge Guided Learning (KGL)~\cite{aakur2022knowledge}, which introduced the notion of open-world egocentric action recognition.
We also create a baseline called ``KGL+CLIP'' by augmenting KGL with CLIP-based grounding by including CLIP's similarity score for establishing semantic correspondences.  
We compare with supervised learning models such as Action Decomposition~\cite{zhang2017first}, IDT~\cite{wang2013action}, and Two-Stream CNN~\cite{simonyan2014two}, with a strong closed-world assumption and a dependency on labeled training data. 
We compare against the zero-shot version of Action Decomposition, which can work under a known world where the final activity labels are known. Not that this is not a fair comparison since it is evaluated under a leave-one-action-out zero-shot learning setting trained on examples from the corresponding dataset for ``seen'' actions. We report it for completeness. 
We also compare against large vision-language models, such as EGO-VLP~\cite{lin2022egocentric}, HierVL~\cite{ashutosh2023hiervl}, and LAVILA~\cite{zhao2023lavila} in both zero-shot and open-world settings (by prompting with all possible verb+noun combinations). 

\begin{table}[t]
    \centering
    \begin{tabular}{|c|c|c|c|c|}
    \toprule
         \textbf{Approach} & \textbf{VLM?} & \textbf{Action} & \textbf{Object} & \textbf{Activity}\\
    \toprule
    Random & \ding{55} & 1.03 & 0.33 & 0.68\\
    KGL~\cite{aakur2022knowledge} & \ding{55} & 3.89 & 2.56 & 3.23\\
    KGL+CLIP~\cite{aakur2022knowledge} & \ding{55} & \underline{5.32} & \underline{4.67} & \underline{4.99}\\
    ALGO (Ours) & \ding{55} & \textbf{10.21} & \textbf{6.76} & \textbf{8.48}\\
    \midrule
    EgoVLP~\cite{lin2022egocentric} & \ding{51} & 10.77 & 19.51 & 15.14\\
    LaViLa~\cite{zhao2023lavila} & \ding{51} & 11.16 & \textbf{23.25} & \underline{17.21}\\
    ALGO+LaViLa & \ding{51} & \textbf{12.54} & \underline{22.84} & \textbf{17.69}\\
    \bottomrule
    \end{tabular}
    \caption{Evaluation on the EPIC-Kitchens-100 dataset. VLM: Vision-Language pre-training on egocentric data. Accuracy for actions, objects, and activity are reported.}
    \label{tab:ek100}
    \vspace{-5mm}
\end{table}
\subsection{Open World Activity Recognition}\label{sec:openWorld}
Table~\ref{tab:gaze_results} summarizes the evaluation results under the open-world inference setting. Top-1 prediction results are reported for all approaches. 
As can be seen, CLIP-based grounding significantly improves the performance of object recognition for KGL, as opposed to the originally proposed, prior-only correspondence function. 
However, our neuro-symbolic grounding mechanism (Section~\ref{sec:NSG}) improves it further, achieving an object recognition performance of $13.07\%$ on Gaze and $26.23\%$ on Gaze Plus. 
It is interesting to note that na\"ively adding CLIP as a mechanism for grounding objects, while effective, does not provide significant gains in the overall action recognition performance (an average of $2\%$ across Gaze and Gaze Plus). We attribute it to the fact that the inherent camera motion in egocentric videos introduces occlusions and visual variations that make it hard to recognize actions consistently. Evidence-based grounding, as proposed in ALGO, makes it more robust to such changes and improves object and action recognition performance. 
Similarly, the posterior-based action refinement module (Section~\ref{sec:posterior}) helps achieve a top-1 action recognition performance of $17.05\%$ on Gaze and $11.44\%$ on Gaze Plus, outperforming KGL ($8.04\%$ and $6.73\%$). 
Adding action priors from LaViLa ($\phi(p(g^a_k | I_t))$ in Equation~\ref{eqn:configEnergy}) allows us to improve the performance further, as indicated by ALGO+LaViLa. We see that LaVila's performance is consistently improved on all metrics. Note that even without the action priors, we outperform LaViLa on GTEA Gaze and offer competitive performance on GazePlus without pre-training on egocentric videos. 

We also evaluate our approach on the Epic-Kitchens-100 dataset, a larger-scale dataset with a significantly higher number of concepts (actions, verbs, and activities). Table~\ref{tab:ek100} summarizes the results. We significantly outperform non-VLM models while offering competitive performance to the VLM-based models. We see that even without \textit{any video-based training data}, we achieve an action accuracy of $10.21\%$ and object accuracy of $6.76\%$, indicating that we can learn affordance-based relationships for discovering and grounding novel actions in egocentric data. Adding action priors from LaViLa further improves the performance. Interestingly, the action (verb) prediction performance of both LaViLa is improved, although it is at the cost of reduced object accuracy. 
Note that the predictions are not separate for verbs and nouns but are computed from the predicted activity. These are remarkable results, considering that the search space is open, i.e., the verb+noun combination is unknown and can be large (380 for Gaze, 405 for Gaze Plus, and 29100 on EPIC-Kitchens). 

To evaluate the generalization capabilities of ALGO,
we presented videos with unseen actions, i.e., actions not in the original training domain, and an open search action space, i.e., not derived from the dataset annotations. We prompted GPT-4~\cite{bubeck2023sparks} using the ChatGPT interface to provide $100$ everyday actions and objects that can be performed in the kitchen to construct our open-world search space and evaluated on GTEA Gaze and GazePlus. 
On unseen actions and unknown search space, the performance was competitive, achieving an accuracy of $9.87\%$ on Gaze and $8.45\%$ on Gaze Plus. 
Please see supplementary for more details and results from the generalization studies. 
These results are encouraging, as they significantly reduce the gap between closed-world learning (supervised), known-world learning (zero-shot), and open-world learning. 

\vspace{-5mm}
\subsubsection{Extension to Zero-Shot Egocentric Activity Recognition}\label{sec:ZSL}
Open-world learning involves the combinatorial search over the different, plausible compositions of elementary concepts. In activity recognition, this involves discovering the action-object (verb-noun) combinations that make up an activity. 
However, in many applications, such as zero-shot recognition, the search space is known, and there is a need to predict pre-specified labels.
To compare our approach with such foundation models, we evaluate ALGO on the Charades-Ego dataset and summarize the results in Table~\ref{tab:ZSL}. We consider the top-10 interpretations made for each clip and perform a nearest neighbor search using ConceptNet Numberbatch embedding to the set of ground-truth labels and pick the one with the least distance. It provides a simple yet effective mechanism to extend our approach to zero-shot settings.
We achieve an mAP score of $16.8\%$ using an S3D~\cite{xie2018rethinking} model pre-trained on Kinetics-400~\cite{kay2017kinetics} and an S3D-G~\cite{miech2020end} model pre-trained on Howto100M~\cite{miech2019howto100m}. This significantly outperforms a comparable TimeSFormer~\cite{bertasius2021space} model pre-trained with a vision-language alignment objective function and provides competitive performance to state-of-the-art vision-language models with significantly lower training requirements - both data and time. 
We observe a similar performance in the Gaze and GazePlus datasets as shown in Table~\ref{tab:gaze_results}. 
We obtain 27.34\% on Gaze and 16.69\% on Gaze Plus, performing competitively with the zero-shot approaches. 
These results are obtained without large amounts of paired text-video pairs and a simple visual-semantic grounding approach. Note that the performance for zero-shot ActionDecomposition is reported for leave-one-class cross-validation, while our approach treats \textit{all classes as unseen classes}. 

\begin{table*}[t]
    \centering
    \resizebox{0.99\textwidth}{!}{
    \begin{tabular}{|c|c|c|c|c|c|c|}
    \toprule
    \multirow{2}{*}{\textbf{Approach}} & \multirow{2}{*}{\textbf{Visual Backbone}} & \multirow{2}{*}{\textbf{Pre-Training?}} & \multicolumn{3}{|c|}{\textbf{Pre-Training Data}}  & \multirow{2}{*}{\textbf{mAP}}\\
    \cline{4-6}
     &  & &  \textbf{Ego?}  & \textbf{Source} &  \textbf{Size} & \\
    \toprule
    EGO-VLP w/o EgoNCE~\cite{lin2022egocentric} & TimeSformer ~\cite{bertasius2021space} & VisLang & \ding{55} & Howto100M \cite{miech2019howto100m} & 136M & 9.2\\
    EGO-VLP w/o EgoNCE ~\cite{lin2022egocentric}& TimeSformer~\cite{bertasius2021space} & VisLang & \ding{55} & CC3M+WebVid-2m & 5.5M & 20.9\\
    EGO-VLP + EgoNCE ~\cite{lin2022egocentric}& TimeSformer~\cite{bertasius2021space} & VisLang & \ding{51} & EgoClip \cite{lin2022egocentric} & 3.8M & 23.6\\
    HierVL ~\cite{ashutosh2023hiervl}& FrozenInTime~\cite{bain2021frozen} & VisLang & \ding{51} & EgoClip \cite{lin2022egocentric} & 3.8M & \underline{26.0}\\
    LAVILA ~\cite{zhao2023lavila}& TimeSformer~\cite{bertasius2021space} & VisLang & \ding{51} & Ego4D~\cite{grauman2022ego4d} & 4M & \textbf{26.8}\\
    \midrule
    ALGO (Ours) & S3D-G ~\cite{miech2019howto100m} & Vision Only & \ding{55} & Howto100M~\cite{miech2019howto100m} & 136M & \textbf{17.3}\\
    ALGO (Ours) & S3D ~\cite{xie2018rethinking} & Vision Only & \ding{55} & Kinetics-400 \cite{kay2017kinetics} & 240K & \underline{16.8}\\
    \bottomrule
    \end{tabular}
    }
    \caption{Evaluation of ALGO under \textbf{zero-shot} learning settings on Charades-Ego where the search space is constrained to ground truth activity semantics. VisLang: Vision Language Pre-Training. }
    \label{tab:ZSL}
    \vspace{-6mm}
\end{table*}


\begin{figure*}[t]
    \centering
    \resizebox{\textwidth}{!}{
    \begin{tabular}{cccc}
     \includegraphics[width=0.5\textwidth]{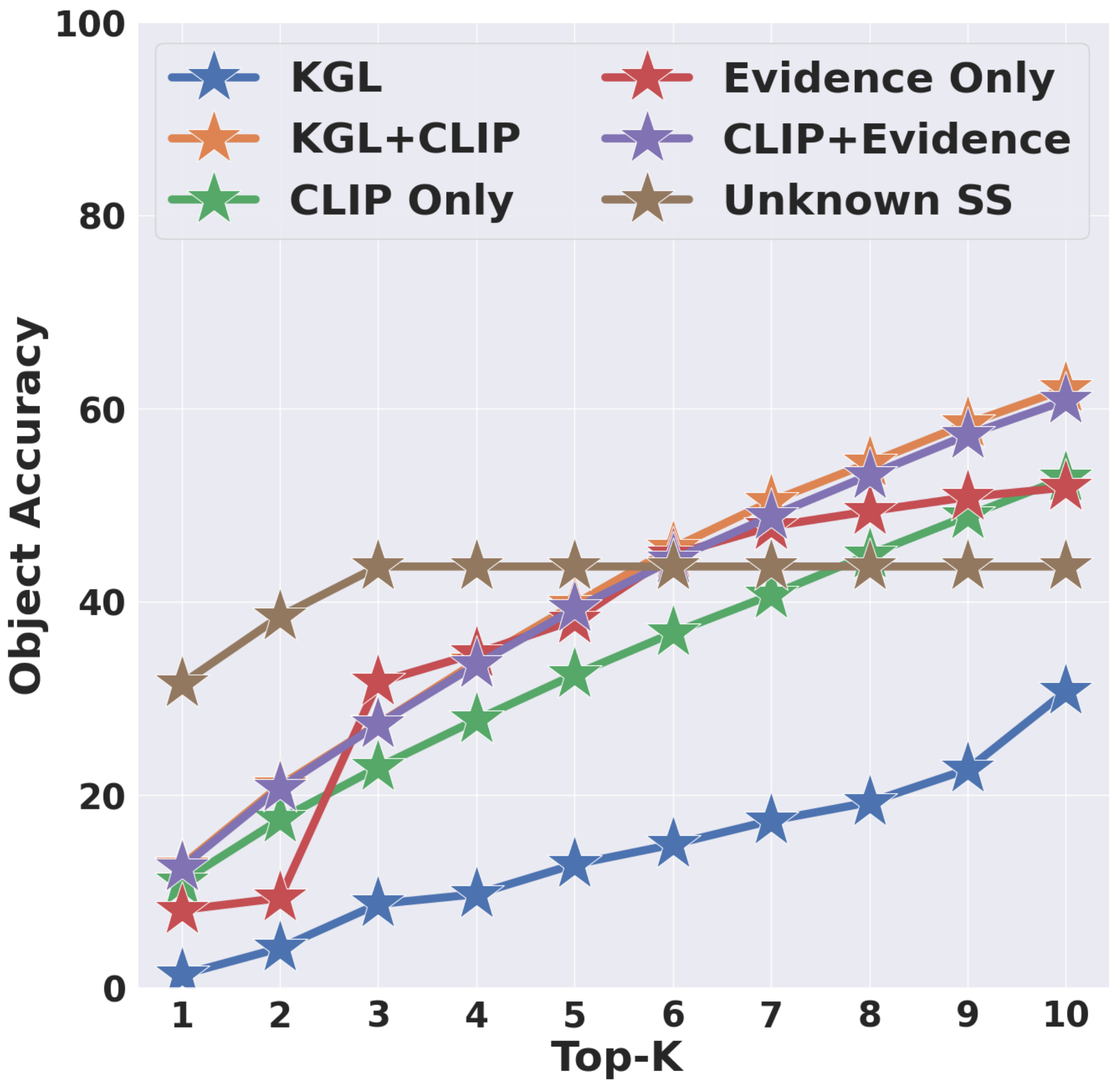}  & \includegraphics[width=0.5\textwidth]{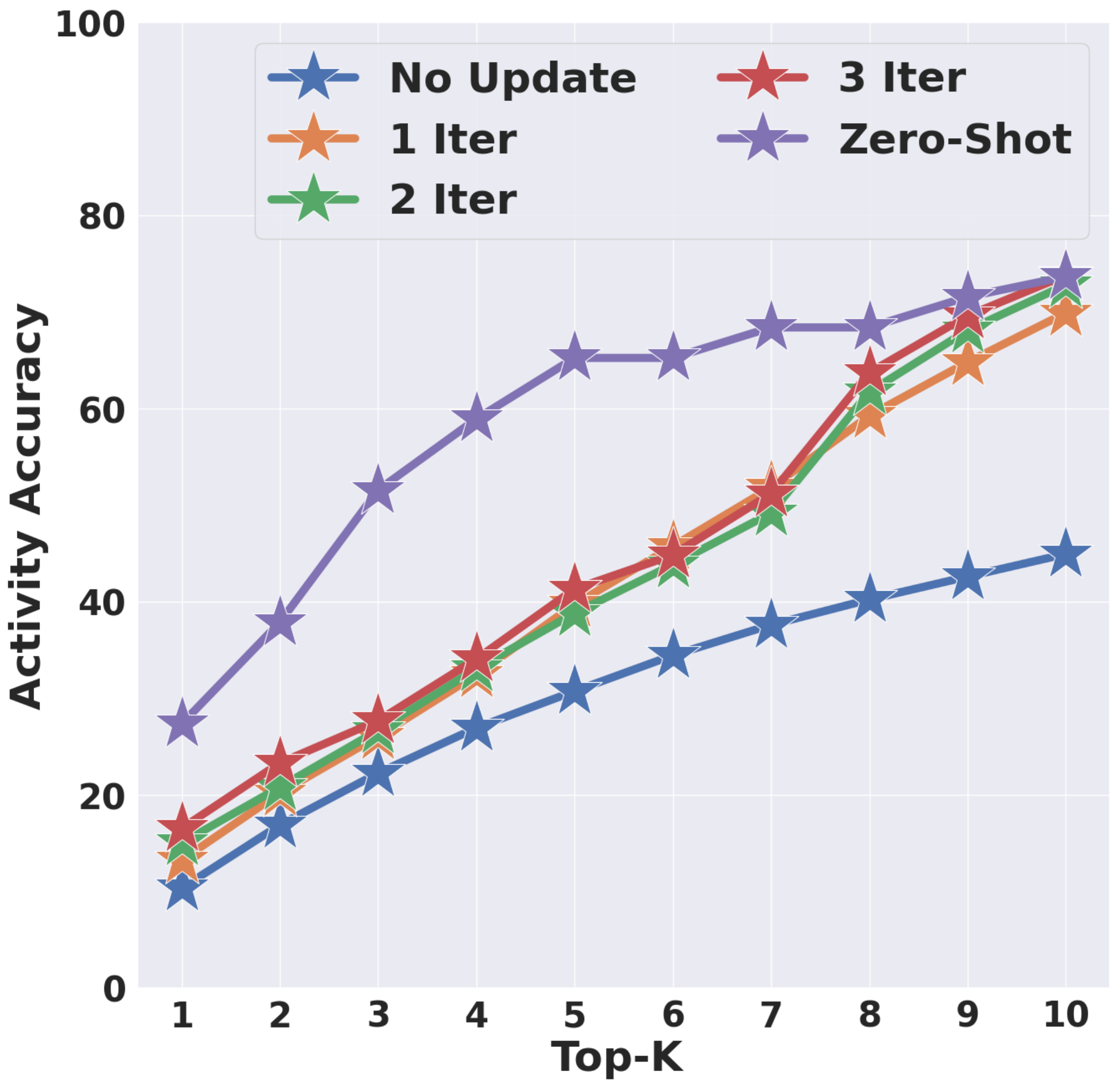} & 
     \includegraphics[width=0.5\textwidth]{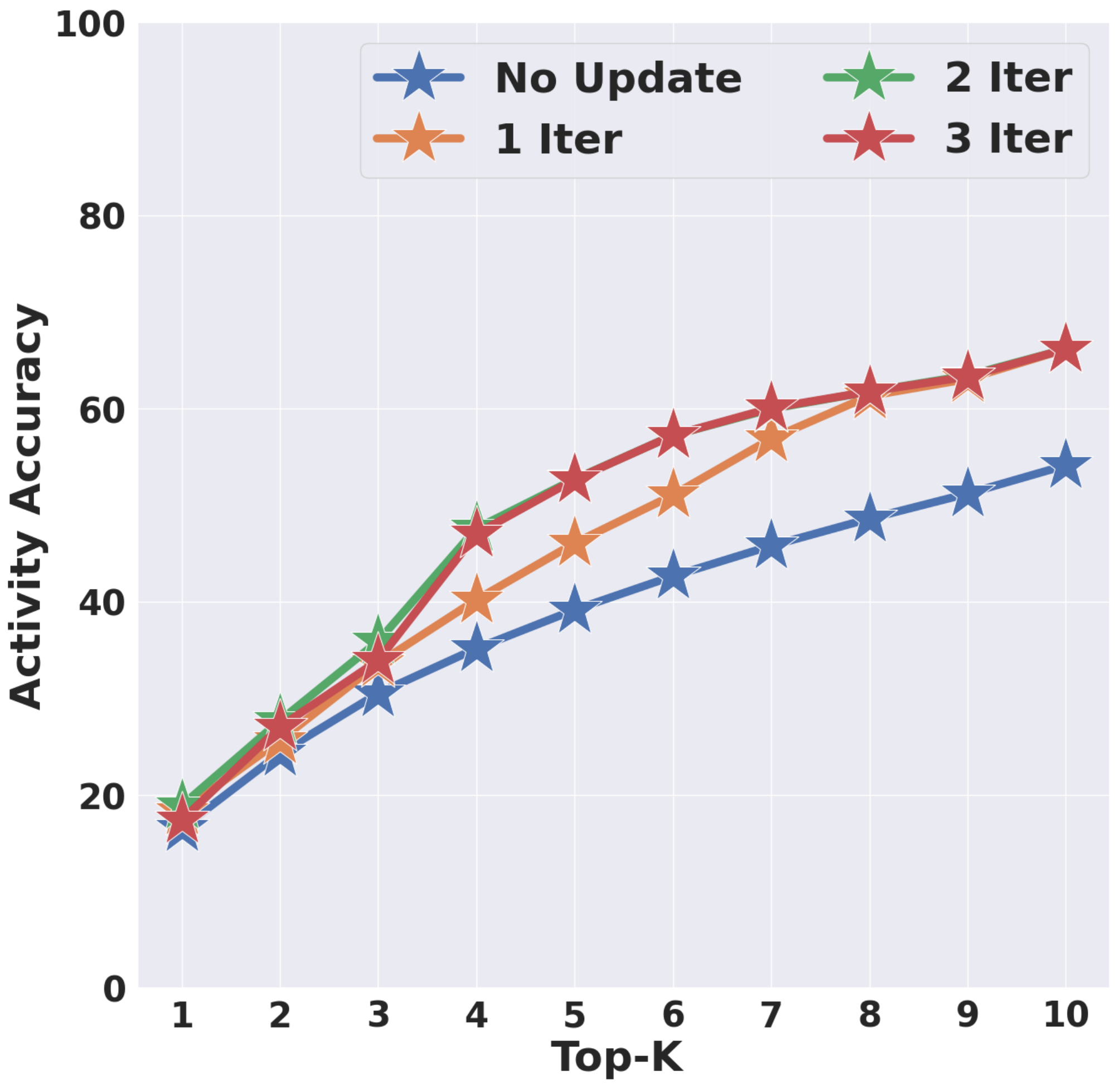} & 
     \includegraphics[width=0.5\textwidth]{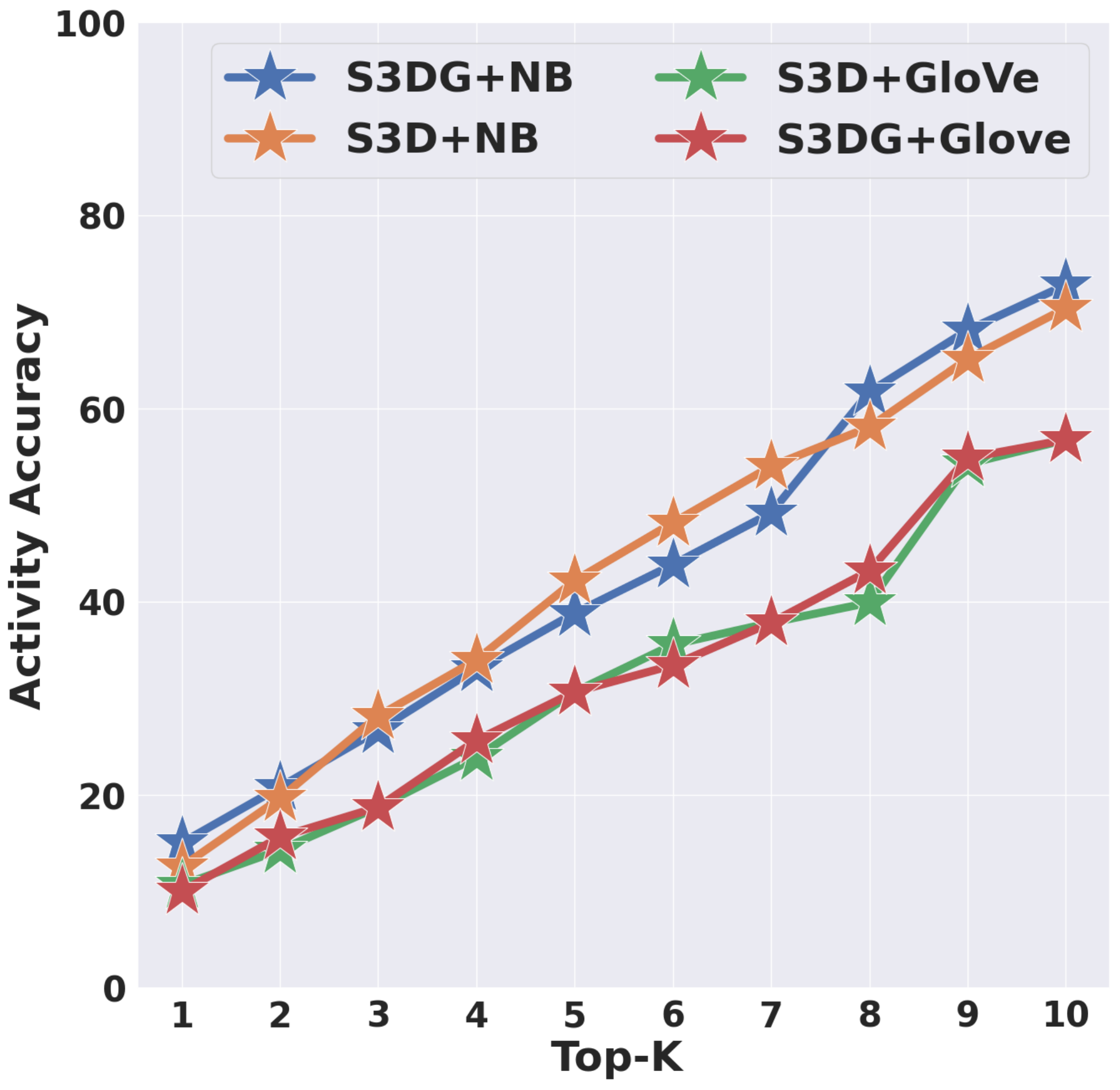}\\
     (a) & (b) & (c) & (d)\\
    \end{tabular}
    }
    \caption{\textbf{Ablation studies} showing the impact of (a) the quality of object grounding techniques, (b) posterior-based action refinement, (c) iterative action refinement on generalization capabilities, and (d) the choice of visual and semantic representations.}
    \label{fig:ablation}
    \vspace{-6mm}
\end{figure*}
\subsection{Ablation Studies}\label{sec:ablation}
We systematically examine the impact of the individual components on the overall performance of the approach. 
Specifically, we focus on three aspects: (i) the impact of object grounding, (ii) the impact of posterior-based action refinement, and (iii) the generalization of learned models with refinement, and (iv) the choice of visual and semantic features. 
We experiment on the GTEA Gaze dataset and present results in Figure~\ref{fig:ablation} and discuss the results below.

\textbf{Quality and Impact of Object Grounding.} First, we evaluate the object recognition performance of different object grounding techniques and present results in Figure~\ref{fig:ablation}(a). We consider 5 different techniques: the prior-based approach proposed in KGL, updating the prior with CLIP-based likelihood (KGL+CLIP), n\"aively using CLIP to recognize the object in the gaze-based ROI (CLIP Only), the proposed evidence-based object grounding (CLIP+Evidence), and using evidence only without checking object-level likelihood (Evidence Only). As can be seen, using CLIP improves performance significantly across the different approaches while using evidence provides gains over the n\"aive CLIP Only method. KGL+CLIP and the proposed CLIP+Evidence approaches perform similarly, with KGL+CLIP being slightly better when considering more than the top-5 recognized objects. 
However, this does not always transfer to better action recognition because the object probabilities are much closer in KGL+CLIP than in the proposed CLIP+Evidence. 
We also evaluated the performance of CLIP+Evidence on an unknown search space by prompting GPT-4 to provide a list of $100$ objects commonly found in the kitchen. The Top-3 performance is excellent, reaching 45\% before saturating, which is remarkable considering that the \textit{unknown} search space with possibly unseen objects. 
We anticipate using visual commonsense priors, such as from scene graphs~\cite{kundu2023ggt}, can help disambiguate between visually similar objects. 

\textbf{Impact of Posterior-based Action Refinement.} One of the major contributions of ALGO is the use of continuous posterior-based action refinement, where the energy of the action generator is refined based on an updated prior from the visual-semantic action grounding to improve the activity recognition performance. 
One key question is how many iterations of such refinements are ideal before overfitting occurs. 
Figure~\ref{fig:ablation}(b) visualizes the activity recognition performance with different levels of iteration, along with the results of a constrained search space (zero-shot) approach. As can be seen, the first two iterations significantly improved the performance, while the third iteration provided very negligible improvement, which provided indications of overfitting. 
Constraining the search space in the zero-shot setting significantly improves the performance. 

\textbf{Generalization of Visual-Semantic Action Grounding.} To evaluate the impact of the posterior-based refinement on the generalization capabilities, we evaluated the trained models, at different iterations, on the GTEA Gaze Plus dataset. As can be seen from Figure~\ref{fig:ablation}, each iteration improves the performance of the model before the performance starts to stagnate (at the third iteration). These results indicate that while iterative refinement is useful, it can lead to overfitting to the domain-specific semantics and can hurt the generalization capabilities of the approach. To this end, we keep the termination criteria for the iterative posterior-based refinement based on the generalization performance of the action grounding model on unseen actions. 

\textbf{Impact of Visual-Semantic Features.} Finally, we evaluate ALGO with different visual and semantic features and visualize the results in Figure~\ref{fig:ablation} (d). We see that the use of ConceptNet Numberbatch (NB) considerably improves the performance of the approach as opposed to using GloVe embeddings~\cite{pennington2014glove}. The choice of visual features (S3DG vs. S3D) does not impact the performance much. We hypothesize that the NB's ability to capture semantic attributes~\cite{speer-lowry-duda-2018-luminoso} allows it to generalize better than GloVe. 
We anticipate custom training of embedding vectors using contextualized, ConceptNet-based pattern theory interpretations could lead to better generalization capabilities. 

\subsection{Generalization of Learned Actions to Unknown Vocabulary}\label{sec:generalization}
To measure the generalization capability of the approach to unknown actions, we use the word similarity score (denoted as NB-WS) to measure the semantic similarity between the predicted and ground truth actions. NB-WS has demonstrated the ability to capture attribute-based representations when computing similarity ~\cite{speer-lowry-duda-2018-luminoso}.
\begin{table*}[t]
    \centering
    \resizebox{0.95\textwidth}{!}{
    \begin{tabular}{|c|c|c|c|c|c|c|c|}
    \toprule
    \multicolumn{2}{|c|}{\textbf{Training Data}} & \multicolumn{2}{|c|}{\textbf{Evaluation Data}} & \multirow{2}{*}{\textbf{Unknown}} & \multirow{2}{*}{\textbf{Search}} & \multirow{2}{*}{\textbf{Accuracy}} & \multirow{2}{*}{\textbf{NB-WS}}\\
    \cmidrule{1-4}
    \textbf{Dataset} & \textbf{\# Verbs} & \textbf{Dataset} & \textbf{\# Verbs} & \textbf{Verbs?} & \textbf{Space} & & \\
    \toprule
    Gaze & 10 & Gaze & 10 & \ding{55} & K & 14.11 & 27.24 \\
    Gaze Plus & 15 & Gaze Plus & 15 & \ding{55} & K & 11.44 & 24.45 \\
    Charades-Ego & 33 & Charades-Ego & 33 & \ding{55} & K & 11.92 & 36.02 \\
    \midrule
    \midrule
    Gaze & 10 & Charades-Ego & 33 & \ding{51} & K & 13.55 & 34.83 \\
    Gaze Plus & 15 & Charades-Ego & 33 & \ding{51} & K & 10.24 & 31.11 \\
    \midrule
    Gaze Plus & 15 & Gaze & 10 & \ding{51} & K & 5.27 & 29.68 \\
    Charades-Ego & 33 & Gaze & 10 & \ding{51} & K & 10.17 & 32.65 \\
    \midrule
    Gaze & 10 & Gaze Plus & 15 & \ding{51} & K & 10.37 & 23.55 \\
    Charades-Ego & 33 & Gaze Plus & 15 & \ding{51} & K & 11.22 & 24.25 \\
    \midrule
    \midrule
    Gaze & 10 & Gaze & 10 & \ding{51} & U & 9.87 & 14.51 \\
    Gaze Plus & 15 & Gaze Plus & 15 & \ding{51} & U & 8.45 & 11.78 \\
    \bottomrule
    \end{tabular}
    }
    \caption{\textbf{Generalization studies} to analyze the performance of the action (verb) recognition models learned in an open-world setting. The models are trained in one domain and evaluated in another, containing possible unknown and unseen actions. NB-WS: ConceptNet Numberbatch Word Similarity}
    \vspace{-2mm}
    \label{tab:generalization}
\end{table*}
We evaluate ALGO's ability to recognize actions from out of its training distribution by presenting videos from datasets with unseen actions and an unknown search space. Specifically, we refer to actions not in the original training domains as ``unseen'' actions, following convention from zero-shot learning. Similarly, in an unknown search space, i.e., \textit{completely open world inference}, the search space is not pre-specified but inferred from general-purpose knowledge sources. For these experiments, we prompted GPT-4~\cite{bubeck2023sparks} using the ChatGPT interface to provide $100$ everyday actions that can be performed in the kitchen to construct our search space. The results are summarized in Table~\ref{tab:generalization}, where we present the verb accuracy and the ConceptNet Numberbatch Word Similarity (NB-WS) score. ALGO generalizes consistently across datasets. Of particular interest is the generalization from Gaze and Gaze Plus to Charades-Ego, where there is a significantly higher number of unseen and unknown actions. Models trained on GTEA Gaze, which has more variation in camera quality and actions, generalize better than those from Gaze Plus. 
With unseen actions and unknown search space, the performance was competitive, achieving an accuracy of $9.87\%$ on Gaze and $8.45\%$ on Gaze Plus. NB-WS was higher, indicating better agreement with the ground truth, i.e., the predicted verbs were similar to the ground truth. 
While there is room for improvement, these results present a significant step towards truly open-world learning without any constraints. 

\section{Discussion, Limitations, and Future Work}
In this work, we proposed ALGO, a neuro-symbolic framework for open-world egocentric activity recognition that aims to learn novel action and activity classes without explicit supervision. By grounding objects and using an object-centered, knowledge-based approach to activity inference, we reduce the need for labeled data to learn semantic associations among elementary concepts. 
We demonstrate that the open-world learning paradigm is an effective inference mechanism to distill commonsense knowledge from symbolic knowledge bases for grounded action understanding. While showing competitive performance, there are two key limitations: (i) it is restricted to ego-centric videos due to the need to navigate clutter by using human attention as a contextual cue for object grounding, and (ii) it requires a knowledge base such as ConceptNet to learn associations between actions and objects and hence is restricted to its vocabulary. 
While we demonstrated its performance on an unknown search space, much work remains to effectively build a search space (both action and object) to move towards a truly open-world learning paradigm. We aim to explore the use of attention-based mechanisms~\cite{mounir2023towards,aakur2022actor} to extend the framework to third-person videos and using abductive reasoning~\cite{aakur2023leveraging,Zellers_2019_CVPR} and neural knowledgebase completion approaches~\cite{bosselut2019comet} to integrate visual commonsense into the reasoning while moving beyond the vocabulary encoded in symbolic knowledgebases. 

\textbf{Acknowledgements.} This research was supported in part by the US National Science Foundation grants IIS 2143150, and IIS 1955230. We thank Dr. Anuj Srivastava (FSU) and Dr. Sudeep Sarkar (USF) for their thoughtful feedback during the discussion about the problem formulation and experimental analysis phase of the project.

\bibliographystyle{splncs04}
\bibliography{main}

\begin{thebibliography}{10}
\providecommand{\url}[1]{\texttt{#1}}
\providecommand{\urlprefix}{URL }
\providecommand{\doi}[1]{https://doi.org/#1}

\bibitem{aakur2022actor}
Aakur, S., Sarkar, S.: Actor-centered representations for action localization in streaming videos. In: Computer Vision--ECCV 2022: 17th European Conference, Tel Aviv, Israel, October 23--27, 2022, Proceedings, Part XXXVIII. pp. 70--87. Springer (2022)

\bibitem{aakur2019generating}
Aakur, S., de~Souza, F., Sarkar, S.: Generating open world descriptions of video using common sense knowledge in a pattern theory framework. Quarterly of Applied Mathematics  \textbf{77}(2),  323--356 (2019)

\bibitem{aakur2021unsupervised}
Aakur, S.N., Bagavathi, A.: Unsupervised gaze prediction in egocentric videos by energy-based surprise modeling. In: International Joint Conference on Computer Vision, Imaging and Computer Graphics Theory and Applications (2021)

\bibitem{aakur2022knowledge}
Aakur, S.N., Kundu, S., Gunti, N.: Knowledge guided learning: Open world egocentric action recognition with zero supervision. Pattern Recognition Letters  \textbf{156},  38--45 (2022)

\bibitem{aakur2023leveraging}
Aakur, S.N., Sarkar, S.: Leveraging symbolic knowledge bases for commonsense natural language inference using pattern theory. IEEE Transactions on Pattern Analysis and Machine Intelligence  (2023)

\bibitem{ashutosh2023hiervl}
Ashutosh, K., Girdhar, R., Torresani, L., Grauman, K.: Hiervl: Learning hierarchical video-language embeddings (2023)

\bibitem{bain2021frozen}
Bain, M., Nagrani, A., Varol, G., Zisserman, A.: Frozen in time: A joint video and image encoder for end-to-end retrieval. In: Proceedings of the IEEE/CVF International Conference on Computer Vision. pp. 1728--1738 (2021)

\bibitem{bertasius2021space}
Bertasius, G., Wang, H., Torresani, L.: Is space-time attention all you need for video understanding? In: International Conference on Machine Learning. pp. 813--824. PMLR (2021)

\bibitem{bommasani2021opportunities}
Bommasani, R., Hudson, D.A., Adeli, E., Altman, R., Arora, S., von Arx, S., Bernstein, M.S., Bohg, J., Bosselut, A., Brunskill, E., et~al.: On the opportunities and risks of foundation models. arXiv preprint arXiv:2108.07258  (2021)

\bibitem{bosselut2019comet}
Bosselut, A., Rashkin, H., Sap, M., Malaviya, C., Celikyilmaz, A., Choi, Y.: Comet: Commonsense transformers for automatic knowledge graph construction. In: Proceedings of the 57th Annual Meeting of the Association for Computational Linguistics. pp. 4762--4779 (2019)

\bibitem{brown2020language}
Brown, T., Mann, B., Ryder, N., Subbiah, M., Kaplan, J.D., Dhariwal, P., Neelakantan, A., Shyam, P., Sastry, G., Askell, A., et~al.: Language models are few-shot learners. Advances in Neural Information Processing Systems  \textbf{33},  1877--1901 (2020)

\bibitem{bubeck2023sparks}
Bubeck, S., Chandrasekaran, V., Eldan, R., Gehrke, J., Horvitz, E., Kamar, E., Lee, P., Lee, Y.T., Li, Y., Lundberg, S., et~al.: Sparks of artificial general intelligence: Early experiments with gpt-4. arXiv preprint arXiv:2303.12712  (2023)

\bibitem{chen2020simple}
Chen, T., Kornblith, S., Norouzi, M., Hinton, G.: A simple framework for contrastive learning of visual representations. In: International Conference on Machine Learning. pp. 1597--1607. PMLR (2020)

\bibitem{clark2020electra}
Clark, K., Luong, M.T., Le, Q.V., Manning, C.D.: Electra: Pre-training text encoders as discriminators rather than generators. arXiv preprint arXiv:2003.10555  (2020)

\bibitem{Damen2021PAMI}
Damen, D., Doughty, H., Farinella, G.M., Fidler, S., Furnari, A., Kazakos, E., Moltisanti, D., Munro, J., Perrett, T., Price, W., Wray, M.: The epic-kitchens dataset: Collection, challenges and baselines. IEEE Transactions on Pattern Analysis and Machine Intelligence (TPAMI)  \textbf{43}(11),  4125--4141 (2021). \doi{10.1109/TPAMI.2020.2991965}

\bibitem{damen2020epic}
Damen, D., Doughty, H., Farinella, G.M., Fidler, S., Furnari, A., Kazakos, E., Moltisanti, D., Munro, J., Perrett, T., Price, W., et~al.: The epic-kitchens dataset: Collection, challenges and baselines. IEEE Transactions on Pattern Analysis and Machine Intelligence  \textbf{43}(11),  4125--4141 (2020)

\bibitem{Damen2022RESCALING}
Damen, D., Doughty, H., Farinella, G.M., Furnari, A., Ma, J., Kazakos, E., Moltisanti, D., Munro, J., Perrett, T., Price, W., Wray, M.: Rescaling egocentric vision: Collection, pipeline and challenges for epic-kitchens-100. International Journal of Computer Vision (IJCV)  \textbf{130},  33–55 (2022), \url{https://doi.org/10.1007/s11263-021-01531-2}

\bibitem{devlin2018bert}
Devlin, J., Chang, M.W., Lee, K., Toutanova, K.: Bert: Pre-training of deep bidirectional transformers for language understanding. arXiv preprint arXiv:1810.04805  (2018)

\bibitem{dong2022open}
Dong, N., Zhang, Y., Ding, M., Lee, G.H.: Open world detr: transformer based open world object detection. arXiv preprint arXiv:2212.02969  (2022)

\bibitem{dosovitskiyimage}
Dosovitskiy, A., Beyer, L., Kolesnikov, A., Weissenborn, D., Zhai, X., Unterthiner, T., Dehghani, M., Minderer, M., Heigold, G., Gelly, S., et~al.: An image is worth 16x16 words: Transformers for image recognition at scale. In: International Conference on Learning Representations

\bibitem{du2022learning}
Du, Y., Wei, F., Zhang, Z., Shi, M., Gao, Y., Li, G.: Learning to prompt for open-vocabulary object detection with vision-language model. In: Proceedings of the IEEE/CVF Conference on Computer Vision and Pattern Recognition. pp. 14084--14093 (2022)

\bibitem{fan2019egovqa}
Fan, C.: Egovqa-an egocentric video question answering benchmark dataset. In: Proceedings of the IEEE/CVF International Conference on Computer Vision Workshops. pp.~0--0 (2019)

\bibitem{fathi2012learning}
Fathi, A., Li, Y., Rehg, J.M.: Learning to recognize daily actions using gaze. In: European Conference on Computer Vision. pp. 314--327. Springer (2012)

\bibitem{grauman2022ego4d}
Grauman, K., Westbury, A., Byrne, E., Chavis, Z., Furnari, A., Girdhar, R., Hamburger, J., Jiang, H., Liu, M., Liu, X., et~al.: Ego4d: Around the world in 3,000 hours of egocentric video. In: Proceedings of the IEEE/CVF Conference on Computer Vision and Pattern Recognition. pp. 18995--19012 (2022)

\bibitem{grenander1996elements}
Grenander, U.: Elements of pattern theory. JHU Press (1996)

\bibitem{Gu2021OpenvocabularyOD}
Gu, X., Lin, T.Y., Kuo, W., Cui, Y.: Open-vocabulary object detection via vision and language knowledge distillation. In: International Conference on Learning Representations (2021), \url{https://api.semanticscholar.org/CorpusID:238744187}

\bibitem{han2020megatrack}
Han, S., Liu, B., Cabezas, R., Twigg, C.D., Zhang, P., Petkau, J., Yu, T.H., Tai, C.J., Akbay, M., Wang, Z., et~al.: Megatrack: monochrome egocentric articulated hand-tracking for virtual reality. ACM Transactions on Graphics (ToG)  \textbf{39}(4),  87--1 (2020)

\bibitem{jia2021scaling}
Jia, C., Yang, Y., Xia, Y., Chen, Y.T., Parekh, Z., Pham, H., Le, Q.V., Sung, Y., Li, Z., Duerig, T.: Scaling up visual and vision-language representation learning with noisy text supervision (2021)

\bibitem{NEURIPS2020_94c28dcf}
Jiang, J., Ahn, S.: Generative neurosymbolic machines. In: Larochelle, H., Ranzato, M., Hadsell, R., Balcan, M., Lin, H. (eds.) Advances in Neural Information Processing Systems. vol.~33, pp. 12572--12582. Curran Associates, Inc. (2020), \url{https://proceedings.neurips.cc/paper_files/paper/2020/file/94c28dcfc97557df0df6d1f7222fc384-Paper.pdf}

\bibitem{kay2017kinetics}
Kay, W., Carreira, J., Simonyan, K., Zhang, B., Hillier, C., Vijayanarasimhan, S., Viola, F., Green, T., Back, T., Natsev, P., et~al.: The kinetics human action video dataset. arXiv preprint arXiv:1705.06950  (2017)

\bibitem{khosla2020supervised}
Khosla, P., Teterwak, P., Wang, C., Sarna, A., Tian, Y., Isola, P., Maschinot, A., Liu, C., Krishnan, D.: Supervised contrastive learning. Advances in Neural Information Processing Systems  \textbf{33},  18661--18673 (2020)

\bibitem{kundu2023ggt}
Kundu, S., Aakur, S.N.: Is-ggt: Iterative scene graph generation with generative transformers. In: Proceedings of the IEEE/CVF Conference on Computer Vision and Pattern Recognition. pp. 6292--6301 (2023)

\bibitem{li2019deep}
Li, H., Cai, Y., Zheng, W.S.: Deep dual relation modeling for egocentric interaction recognition. In: Proceedings of the IEEE/CVF Conference on Computer Vision and Pattern Recognition. pp. 7932--7941 (2019)

\bibitem{li2022supervision}
Li, Y., Liang, F., Zhao, L., Cui, Y., Ouyang, W., Shao, J., Yu, F., Yan, J.: Supervision exists everywhere: A data efficient contrastive language-image pre-training paradigm (2022)

\bibitem{li2013learning}
Li, Y., Fathi, A., Rehg, J.M.: Learning to predict gaze in egocentric video. In: Proceedings of the IEEE International Conference on Computer Vision. pp. 3216--3223 (2013)

\bibitem{lin2022egocentric}
Lin, K.Q., Wang, J., Soldan, M., Wray, M., Yan, R., XU, E.Z., Gao, D., Tu, R.C., Zhao, W., Kong, W., et~al.: Egocentric video-language pretraining. Advances in Neural Information Processing Systems  \textbf{35},  7575--7586 (2022)

\bibitem{liu2019roberta}
Liu, Y., Ott, M., Goyal, N., Du, J., Joshi, M., Chen, D., Levy, O., Lewis, M., Zettlemoyer, L., Stoyanov, V.: Roberta: A robustly optimized bert pretraining approach. arXiv preprint arXiv:1907.11692  (2019)

\bibitem{lu2013story}
Lu, Z., Grauman, K.: Story-driven summarization for egocentric video. In: Proceedings of the IEEE conference on Computer Vision and Pattern Recognition. pp. 2714--2721 (2013)

\bibitem{ma2016going}
Ma, M., Fan, H., Kitani, K.M.: Going deeper into first-person activity recognition. In: IEEE/CVF Conference on Computer Vision and Pattern Recognition (CVPR). pp. 1894--1903 (2016)

\bibitem{maguire2008speaking}
Maguire, M.J., Dove, G.O.: Speaking of events: event word learning and event representation. Understanding Events: How Humans See, Represent, and Act on Events pp. 193--218 (2008)

\bibitem{menon2022visual}
Menon, S., Vondrick, C.: Visual classification via description from large language models. arXiv preprint arXiv:2210.07183  (2022)

\bibitem{miech2020end}
Miech, A., Alayrac, J.B., Smaira, L., Laptev, I., Sivic, J., Zisserman, A.: End-to-end learning of visual representations from uncurated instructional videos. In: Proceedings of the IEEE/CVF Conference on Computer Vision and Pattern Recognition. pp. 9879--9889 (2020)

\bibitem{miech2019howto100m}
Miech, A., Zhukov, D., Alayrac, J.B., Tapaswi, M., Laptev, I., Sivic, J.: Howto100m: Learning a text-video embedding by watching hundred million narrated video clips. In: Proceedings of the IEEE/CVF International Conference on Computer Vision. pp. 2630--2640 (2019)

\bibitem{mounir2023towards}
Mounir, R., Shahabaz, A., Gula, R., Theuerkauf, J., Sarkar, S.: Towards automated ethogramming: Cognitively-inspired event segmentation for streaming wildlife video monitoring. International Journal of Computer Vision pp. 1--31 (2023)

\bibitem{nye2021improving}
Nye, M., Tessler, M., Tenenbaum, J., Lake, B.M.: Improving coherence and consistency in neural sequence models with dual-system, neuro-symbolic reasoning. Advances in Neural Information Processing Systems  \textbf{34},  25192--25204 (2021)

\bibitem{pennington2014glove}
Pennington, J., Socher, R., Manning, C.: Glove: Global vectors for word representation. In: Proceedings of the 2014 Conference on Empirical Methods in Natural Language Processing (EMNLP). pp. 1532--1543 (2014)

\bibitem{radford2021learning}
Radford, A., Kim, J.W., Hallacy, C., Ramesh, A., Goh, G., Agarwal, S., Sastry, G., Askell, A., Mishkin, P., Clark, J., et~al.: Learning transferable visual models from natural language supervision. In: International Conference on Machine Learning. pp. 8748--8763. PMLR (2021)

\bibitem{radford2018improving}
Radford, A., Narasimhan, K., Salimans, T., Sutskever, I.: Improving language understanding by generative pre-training  (2018)

\bibitem{radford2019language}
Radford, A., Wu, J., Child, R., Luan, D., Amodei, D., Sutskever, I., et~al.: Language models are unsupervised multitask learners. OpenAI blog  \textbf{1}(8), ~9 (2019)

\bibitem{Ryoo_2015_CVPR}
Ryoo, M.S., Rothrock, B., Matthies, L.: Pooled motion features for first-person videos. In: Proceedings of the IEEE conference on Computer Vision and Pattern Recognition (CVPR) (June 2015)

\bibitem{sigurdsson2018actor}
Sigurdsson, G.A., Gupta, A., Schmid, C., Farhadi, A., Alahari, K.: Actor and observer: Joint modeling of first and third-person videos. In: Proceedings of the IEEE/CVF International Conference on Computer Vision. pp. 7396--7404 (2018)

\bibitem{simonyan2014two}
Simonyan, K., Zisserman, A.: Two-stream convolutional networks for action recognition in videos. Advances in Neural Information Processing Systems  \textbf{27} (2014)

\bibitem{desouza2016pattern}
de~Souza, F.D., Sarkar, S., Srivastava, A., Su, J.: Pattern theory for representation and inference of semantic structures in videos. Pattern Recognition Letters  \textbf{72},  41--51 (2016)

\bibitem{speer2017conceptnet}
Speer, R., Chin, J., Havasi, C.: Conceptnet 5.5: An open multilingual graph of general knowledge. In: Proceedings of the AAAI Conference on Artificial Intelligence. vol.~31 (2017)

\bibitem{speer-lowry-duda-2018-luminoso}
Speer, R., Lowry-Duda, J.: {L}uminoso at {S}em{E}val-2018 task 10: Distinguishing attributes using text corpora and relational knowledge. In: Proceedings of the 12th International Workshop on Semantic Evaluation. pp. 985--989. Association for Computational Linguistics, New Orleans, Louisiana (Jun 2018). \doi{10.18653/v1/S18-1162}, \url{https://aclanthology.org/S18-1162}

\bibitem{Sudhakaran_2019_CVPR}
Sudhakaran, S., Escalera, S., Lanz, O.: Lsta: Long short-term attention for egocentric action recognition. In: Proceedings of the IEEE/CVF Conference on Computer Vision and Pattern Recognition (CVPR) (June 2019)

\bibitem{vaswani2017attention}
Vaswani, A., Shazeer, N., Parmar, N., Uszkoreit, J., Jones, L., Gomez, A.N., Kaiser, {\L}., Polosukhin, I.: Attention is all you need. Advances in Neural Information Processing Systems  \textbf{30} (2017)

\bibitem{wang2013action}
Wang, H., Schmid, C.: Action recognition with improved trajectories. In: IEEE/CVF International Conference on Computer Vision (ICCV). pp. 3551--3558 (2013)

\bibitem{Wang_2021_ICCV}
Wang, X., Zhu, L., Wang, H., Yang, Y.: Interactive prototype learning for egocentric action recognition. In: Proceedings of the IEEE/CVF International Conference on Computer Vision (ICCV). pp. 8168--8177 (October 2021)

\bibitem{NEURIPS2022_3ff48dde}
Wu, T., Tjandrasuwita, M., Wu, Z., Yang, X., Liu, K., Sosic, R., Leskovec, J.: Zeroc: A neuro-symbolic model for zero-shot concept recognition and acquisition at inference time. In: Koyejo, S., Mohamed, S., Agarwal, A., Belgrave, D., Cho, K., Oh, A. (eds.) Advances in Neural Information Processing Systems. vol.~35, pp. 9828--9840. Curran Associates, Inc. (2022), \url{https://proceedings.neurips.cc/paper_files/paper/2022/file/3ff48dde82306fe8f26f3e51dd1054d7-Paper-Conference.pdf}

\bibitem{xie2018rethinking}
Xie, S., Sun, C., Huang, J., Tu, Z., Murphy, K.: Rethinking spatiotemporal feature learning: Speed-accuracy trade-offs in video classification. In: Proceedings of the European Conference on Computer Vision. pp. 305--321 (2018)

\bibitem{yu2022coca}
Yu, J., Wang, Z., Vasudevan, V., Yeung, L., Seyedhosseini, M., Wu, Y.: Coca: Contrastive captioners are image-text foundation models (2022)

\bibitem{Zellers_2019_CVPR}
Zellers, R., Bisk, Y., Farhadi, A., Choi, Y.: From recognition to cognition: Visual commonsense reasoning. In: Proceedings of the IEEE/CVF Conference on Computer Vision and Pattern Recognition (CVPR) (June 2019)

\bibitem{zhang2017first}
Zhang, Y.C., Li, Y., Rehg, J.M.: First-person action decomposition and zero-shot learning. In: IEEE Winter Conference on Applications of Computer Vision (WACV). pp. 121--129 (2017)

\bibitem{zhao2023lavila}
Zhao, Y., Misra, I., Kr{\"a}henb{\"u}hl, P., Girdhar, R.: Learning video representations from large language models. In: CVPR (2023)

\bibitem{Zhou_2016_CVPR}
Zhou, Y., Ni, B., Hong, R., Yang, X., Tian, Q.: Cascaded interactional targeting network for egocentric video analysis. In: Proceedings of the IEEE conference on Computer Vision and Pattern Recognition (CVPR) (June 2016)

\end{thebibliography}

\newpage
\clearpage
\setcounter{page}{1}

\section*{Implementation Details.} 
We use an S3D-G network pre-trained by Miech \textit{et al.}~\cite{miech2019howto100m,miech2020end} on Howto100M~\cite{miech2019howto100m} as our visual feature extraction for visual-semantic action grounding. We use a CLIP model with the ViT-B/32~\cite{dosovitskiyimage} as its backbone network. ConceptNet was used as our source of commonsense knowledge for neuro-symbolic reasoning, and ConceptNet Numberbatch~\cite{speer2017conceptnet} was used as the semantic representation for action grounding. 
The MCMC-based inference from KGL~\cite{aakur2022knowledge} was used as our reasoning mechanism. 
For experiments on Charades-Ego, where gaze information was unavailable, center bias~\cite{li2013learning} was used to approximate gaze locations. 
The mapping function, defined in Section~\ref{sec:temporalGround}, was a 1-layer feedforward network trained with the MSE loss for 100 epochs with a batch size of 256 and learning rate of $10^{-3}$. Generalization errors on unseen actions were used to pick the best model. 
Two iterations of posterior-based action refinement were performed per video. 
Experiments were conducted on a desktop with a 32-core AMD ThreadRipper and an NVIDIA Titan RTX.

\section*{Why temporal smoothing?}
Since we predict frame-level activity interpretations to account for gaze transitions, we first perform temporal smoothing to label the entire video clip before training the mapping function $\psi(g^a_i, f_V)$ to reduce noise in the learning process. For each frame in the video clip, we take the five most common actions predicted at the \textit{activity} level (considering the top-10 predictions) and sum their energies to consolidate activity predictions and account for erroneous predictions. We then repeat the process for the entire clip, i.e., get the top-5 actions based on their frequency of occurrence at the frame level and consolidated energies across frames. These five actions provide targets for the mapping function $\psi(g^a_i, f_V)$, which is then trained with the MSE function. We use the top-5 action labels as targets to limit the effect of frequency bias. 
For example, some actions, such as \textit{clean}, can possess a high affinity to many objects and hence be the most commonly predicted action for a frame. 
Hence, temporal smoothing acts as a regularizer to reduce overfitting by forcing the model to predict the embedding for the top five actions for each video clip. 

\section*{Why posterior-based refinement?}
 Since our predictions are made on a per-frame basis, it does not consider the overall temporal coherence and visual dynamics of the clip. Hence, there can be contradicting predictions for the actions done over time. 
Similarly, when setting the action priors to $1$, we consider all actions equally plausible and do not restrict the action labels through grounding, as done for objects in Section~\ref{sec:NSG}. 
Hence, we iteratively update the action priors for the energy computation to re-rank the interpretations based on the clip-level visual dynamics. 
This prior could be updated to consider predictions from other models, such as EGO-VLP~\cite{lin2022egocentric} through prompting mechanisms similar to our neuro-symbolic object grounding. 
We iteratively refine the activity labels and update the visual-semantic action grounding modules simultaneously by alternating between posterior update and action grounding until the generalization error (i.e., the performance on unseen actions) saturates, which indicates overfitting. 
We empirically verify this in Section~\ref{sec:ablation}, where we observe the impact of this posterior update on activity recognition performance and the resulting loss of generalization.

\section*{Datasets}
The GTEA Gaze dataset consists of 14 subjects performing activities composed of 10 verbs and 38 nouns across 17 videos. 
The gaze information is collected using Tobii eye-tracking glasses at 30 frames per second. 
The Gaze Plus dataset has 27 nouns and 15 verbs from 6 subjects performing 7 meal preparation activities across 37 videos. 
The gaze information is collected at 30 frames per second for both datasets using SMI eye-tracking glasses. 
Charades-Ego contains 7,860 videos containing 157 activities. Following prior work~\cite{lin2022egocentric}, we use the 785 egocentric clips in the test set for evaluation. EpicKitchens-100 is a large dataset comprising several hours of egocentric videos with 300 nouns (objects) and 97 verbs (actions). We use the validation set to evaluate our approach following prior works~\cite{zhao2023lavila}.

\section*{Visualizations of generated interpretations}
\begin{figure*}
    \centering
    \begin{tabular}{cc}
             \includegraphics[width=0.45\columnwidth]{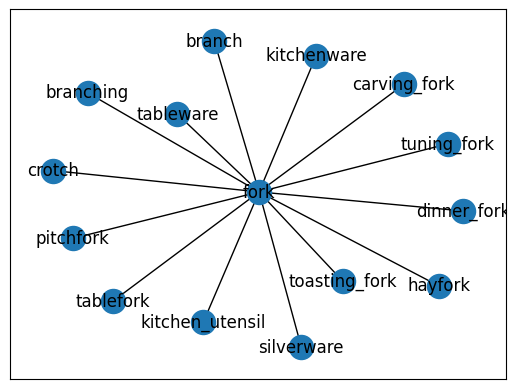} &
             \includegraphics[width=0.45\columnwidth]{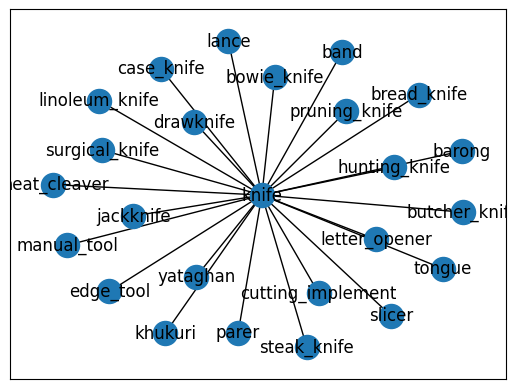} \\
             (a) & (b) \\
             \includegraphics[width=0.45\columnwidth]{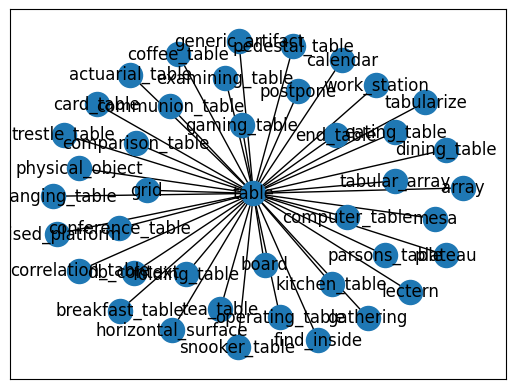} &
             \includegraphics[width=0.45\columnwidth]{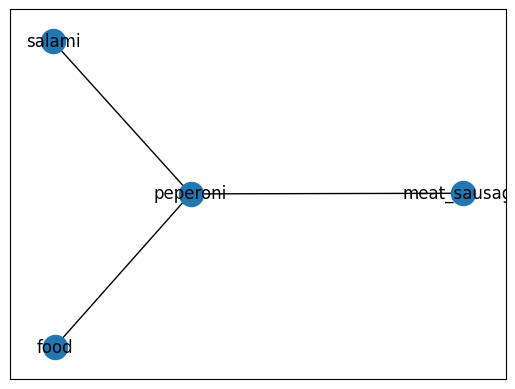}\\
             (c) & (d) \\
             \includegraphics[width=0.45\columnwidth]{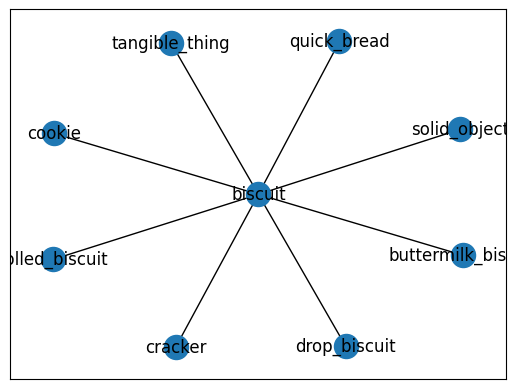} &
             \includegraphics[width=0.45\columnwidth]{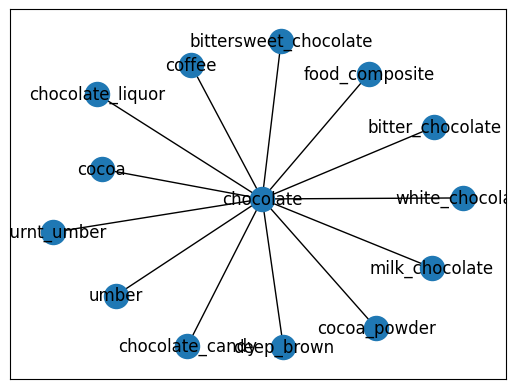}\\
             (e) & (f) \\

    \end{tabular}
    \caption{Visualization of alternative concepts that were tested for grounding concepts in the video such as (a) fork, (b) knife, (c) table, (d) pepperoni, (e) biscuit, and (f) chocolate. These are automatically derived from ConceptNet and have semantic assertions quantifying how they are related.}
    \label{fig:qual_evidence}
\end{figure*}

\textbf{Evidence-based grounding.} Some examples of evidence-based grounding through ConceptNet are shown in Figure~\ref{fig:qual_evidence}. As can be seen, each concept can have multiple evidence generators derived from ConceptNet using its ego-graph and limiting edges to those that express \textit{compositional} properties such as \texttt{IsA, UsedFor, HasProperty} and \texttt{SynonymOf}. 
Using ego-graph helps preserve the contextual information within the semantic locality of the object to filter high-order noise induced by regular k-hop neighborhoods. The derived concepts provide additional context for verifying the presence of a given object in the video by querying CLIP as a noisy object oracle. Note that we do not visualize the edge label to avoid clutter. Each edge is qualified by a semantic assertion and is quantified by a value between -2 and 2 to express its strength. This provides a more explainable representation that enhances the final interpretation generated, as discussed next.

\begin{figure*}
    \centering
    \begin{tabular}{cc}
             \includegraphics[width=0.45\columnwidth]{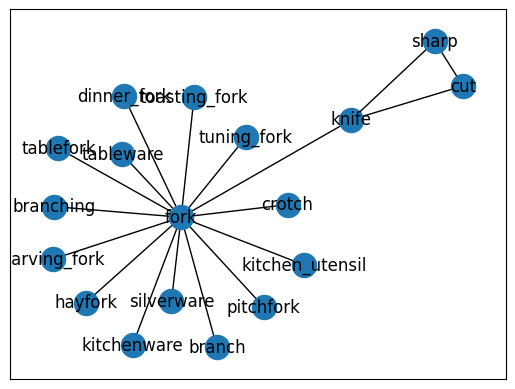} &
             \includegraphics[width=0.45\columnwidth]{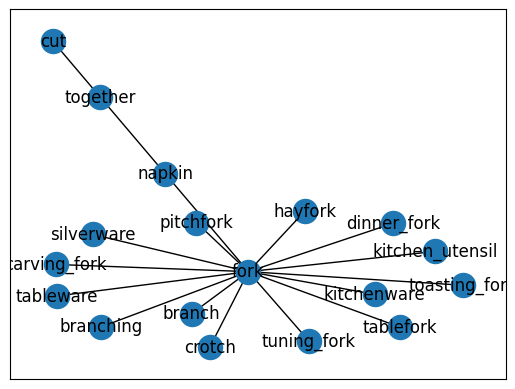} \\
             (a) & (b) \\
             \includegraphics[width=0.45\columnwidth]{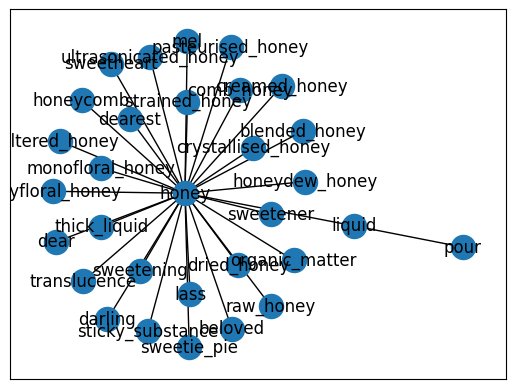} &
             \includegraphics[width=0.45\columnwidth]{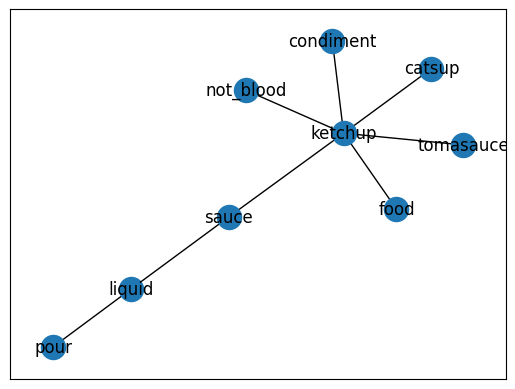}\\
             (c) & (d) \\
             \includegraphics[width=0.45\columnwidth]{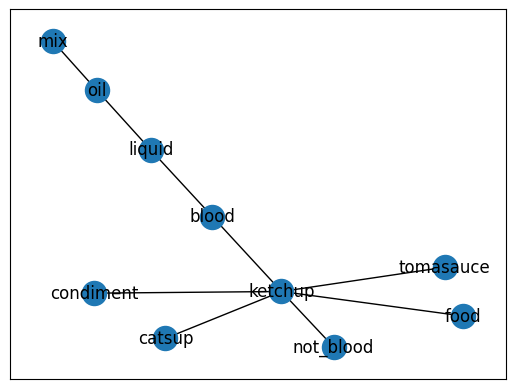} &
             \includegraphics[width=0.45\columnwidth]{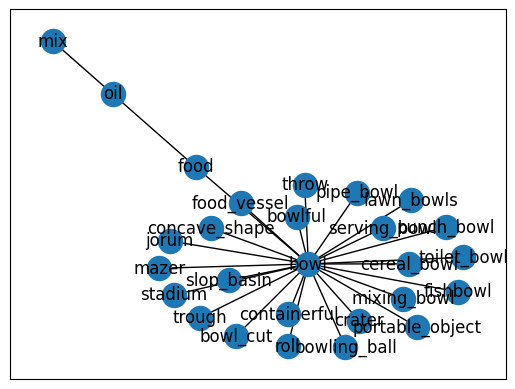}\\
             (e) & (f) \\

    \end{tabular}
    \caption{Visualization of final interpretations for videos containing the activity (a) cut fork (top interpretation), (b) cut fork (second best interpretation), (c) pour honey, (d) pour ketchup, (e) mix ketchup, and (f) mix bowl. These are automatically derived from ConceptNet and have semantic assertions quantifying how they are related.}
    \label{fig:qual_interpretation}
\end{figure*}

\textbf{Semantically rich interpretations.} The final interpretations generated by the approach are shown in Figure~\ref{fig:qual_interpretation}. It can be seen that the final interpretation is semantically rich and has concepts that are not directly in the scene but are compositionally relevant, either through affordance or object-level evidence. We visualize two different interpretations for affordance-based reasoning for the activity ``cut fork'' in (a) and (b), where it can be seen that the approach can generate graphs of varying structures. It also shows the impact of the noise in the knowledge graph that can introduce irrelevant concepts into the reasoning process. We anticipate that having an additional reasoning step can reduce the impact of noise. Interestingly, we see that combinations of similar verbs and nouns such as ``pour honey'' (c) and ``pour ketchup'' (d) result in different ungrounded generators, indicating that the affordance of each concept is used for reasoning, resulting in semantically rich graphical interpretations. We anticipate that learning customized embeddings from these graphs can result in a better grounding of novel compositional concepts such as actions and activities. 

\textbf{Baselines}. Since it is the most comparable, deep learning-only baseline, we choose ActionDecomposition [65] as our primary ZSL baseline for the GTEA datasets. It decouples verb and noun recognition and uses similar feature extractors for noun+verb prediction. However, they only report leave-one-class-out cross-validation, which assumes access to other examples from ``seen'' classes during training, which is not a fair comparison with our approach. 
We do not require any labels during training. 
Given the list of possible objects and actions in the videos, we ground the objects using CLIP, infer plausible actions using affordance-based reasoning, and predict the final activities driven purely by prior knowledge encoded in ConceptNet. 
{Besides CLIP (trained only on objects), ALGO is not trained with any labels. Zero-shot models and ``foundation models'' are trained on considerable amounts of \textit{video} data \textit{with action/activity/object} labels and learn verb+noun associations from these labels. While ``chain-of-thought prompting'' or other decomposition approaches can possibly improve their generalization, \underline{no such models exist.}} We report the supervised/zero-shot/VLMs performance to illustrate that ALGO performs competitively despite not requiring large-scale video pretraining and labels (especially for actions and activities). 
While there is a gap between their performance, note that we do not have access to any data about actions (verbs) or activities, while VLMs and ZS models are trained exclusively on videos with such information. We learn to recognize actions with no labels and show that VLMs can benefit from ALGO (ALGO+LAVILA and ALGO+EGOVLP). 

\textbf{Action Recognition and Metrics}. While activity recognition is the focus of the work, we would like to point out that the action (verb) recognition is also done in an open-world, inferred without labeled data. The performances reported for ``Action'' in Tables 1-3 represent ALGO's ability to infer the verb from contextual information. The top-5 performance of ALGO (obj/action/activity) on Gaze (40.75/34.89/37.82 vs. KGL's 32.39/10.73/18.78) and Gaze+ (42.77/53.88/48.33 vs. KGL's 24.64/37.99/27.53) as well as the exact match activity accuracy (Gaze - ALGO: 1.8, ALGO+LAVILA: 3.5, KGL: 0.3, EGOVLP: 0.9, LAVILA: 2.1, Gaze+ - ALGO: 3.4, ALGO+LAVILA: 8.3, KGL: 1.1, EGOVLP: 4.1, LAVILA: 7.9) show consistent improvements over the baselines across all metrics. While VLMs perform better than ALGO on exact match, which we attribute to their ability to identify the verb correctly, augmenting them with ALGO consistently improves their performance. 

\textbf{Search space and knowledge source}.  
We use ConceptNet as the source of knowledge, over LLM or word embedding approaches, due to its ability to support probabilistic reasoning and an interpretable internal mechanism (see [4]). 
While this does limit its vocabulary, it serves us well for reasoning over action-object affordance and affinity. 
We could replace ConceptNet priors with word embedding techniques such as GloVe or GPT 4, and the framework would still function without any modifications. Preliminary analysis with GloVe embedding on GTEA Gaze results in object/action/activity accuracy of 12.86/14.53/13.70, which outperforms KGL/KGL+CLIP/LAVILA/EGOVLP. More complex embedding could improve this performance at the cost of interpretability. Additionally, we show in Fig 2a that ConceptNet can be augmented with ChatGPT to move beyond its vocabulary. \textbf{The vocabulary of ConceptNet is not the same as the search space for CLIP/LAVILA/EGOVLP.} The search space is the space of prompts one provides to VLMs to select from and is usually pre-defined in zero-shot inference. Open-world inference requires building this search space for VLMs to infer labels. We propose (including prior-driven prompting) to construct such a search space without brute-force search over all combinations of verbs/nouns. 

\textbf{Qualitative Analysis}
Our analysis shows that the performance across the datasets varies and primarily stems from how the verbs and nouns are defined in the dataset. For example, Gaze and Gaze+ have activity labels with less ambiguous verb-noun combinations than EK100. For example, ``clean'', ``put'', and ``take'', common verbs in EK100, can apply to every single object and have very high affinity in ConceptNet, leading to a higher likelihood of prediction. This is one of the failure modes of ALGO and is one of our future research directions.

\end{document}